\documentclass{article}

\usepackage{microtype}
\usepackage{graphicx}
\usepackage{subcaption}
\usepackage{booktabs} 

\usepackage{hyperref}

\usepackage[accepted]{arxiv}

\usepackage{amsmath}
\usepackage{amssymb}
\usepackage{mathtools}
\usepackage{amsthm}

\usepackage[capitalize,noabbrev]{cleveref}

\theoremstyle{plain}

\theoremstyle{definition}

\theoremstyle{remark}

\usepackage[textsize=tiny]{todonotes}

\renewcommand{\paragraph}[1]{\vspace{-0.1em} \noindent \textbf{#1}}
\newcommand{\sparsegpt}[1]{\texttt{SparseGPT}}

\icmltitlerunning{SparseGPT: Massive Language Models Can be Accurately Pruned in One-Shot}

\begin{document}

\twocolumn[
\icmltitle{SparseGPT: Massive Language Models Can be Accurately Pruned in One-Shot}

\icmlsetsymbol{equal}{*}

\begin{icmlauthorlist}
    \icmlauthor{Elias Frantar}{ista}
    \icmlauthor{Dan Alistarh}{ista,nm}
\end{icmlauthorlist}

\icmlaffiliation{ista}{Institute of Science and Technology Austria (ISTA)}
\icmlaffiliation{nm}{Neural Magic Inc}

\icmlcorrespondingauthor{Elias Frantar}{elias.frantar@ist.ac.at}

\vskip 0.3in
]

\printAffiliationsAndNotice{}

\begin{abstract}
We show for the first time that large-scale generative pretrained transformer (GPT) family models can be pruned to at least 50\% sparsity in \emph{one-shot, without any retraining}, at minimal loss of accuracy. 
This is achieved via a new pruning method called \sparsegpt{}, specifically designed to work efficiently and accurately on  massive GPT-family models.
We can execute \sparsegpt{} on the largest available open-source models, OPT-175B and BLOOM-176B, in under 4.5 hours, and can reach 60\% unstructured sparsity with negligible increase in perplexity: remarkably, more than 100 billion weights from these models can be ignored at inference time.
\sparsegpt{} generalizes to semi-structured (2:4 and 4:8) patterns,  and is compatible with weight quantization approaches. The code is available at: \url{https://github.com/IST-DASLab/sparsegpt}.
\end{abstract}

\section{Introduction}

Large Language Models (LLMs) from the Generative Pretrained Transformer (GPT) family have shown remarkable performance on a wide range of tasks, but are difficult to deploy because of their massive size and computational costs.  
For illustration, the top-performing GPT-175B models have 175 billion parameters, which total at least 320GB (counting multiples of 1024) of storage in half-precision (FP16) format, leading it to require at least five A100 GPUs with 80GB of memory each for inference. 
It is therefore natural that there has been significant interest in reducing these costs via \emph{model compression}. 
To date, virtually all existing GPT compression approaches have focused on \emph{quantization}~\cite{dettmers2022llm, yao2022zeroquant, xiao2022smoothquant, frantar2022gptq}, that is, reducing the precision of the model's numerical representation. 

A complementary approach for compression is \emph{pruning}, which removes network elements, 
from individual weights (unstructured pruning) to higher-granularity structures such as rows/columns of the weight matrices (structured pruning). 
Pruning has a long history~\cite{lecun1990optimal, hassibi1993optimal}, and has been applied successfully in the case of vision and smaller-scale language models~\cite{hoefler2021sparsity}. 
Yet, the best-performing pruning methods require \emph{extensive retraining} of the model to recover accuracy. 
In turn, this is extremely expensive for GPT-scale models. 
While some accurate \emph{one-shot} pruning methods exist~\cite{hubara2021accelerated, frantar2022obc}, compressing the model without retraining, unfortunately even they become very expensive when applied to models with billions of parameters. 
Thus, to date, there is essentially no work on accurate pruning of billion-parameter models.

\begin{figure*}[h]
    \begin{minipage}[c]{.49\textwidth}
        \centering
        \includegraphics[width=.8\linewidth]{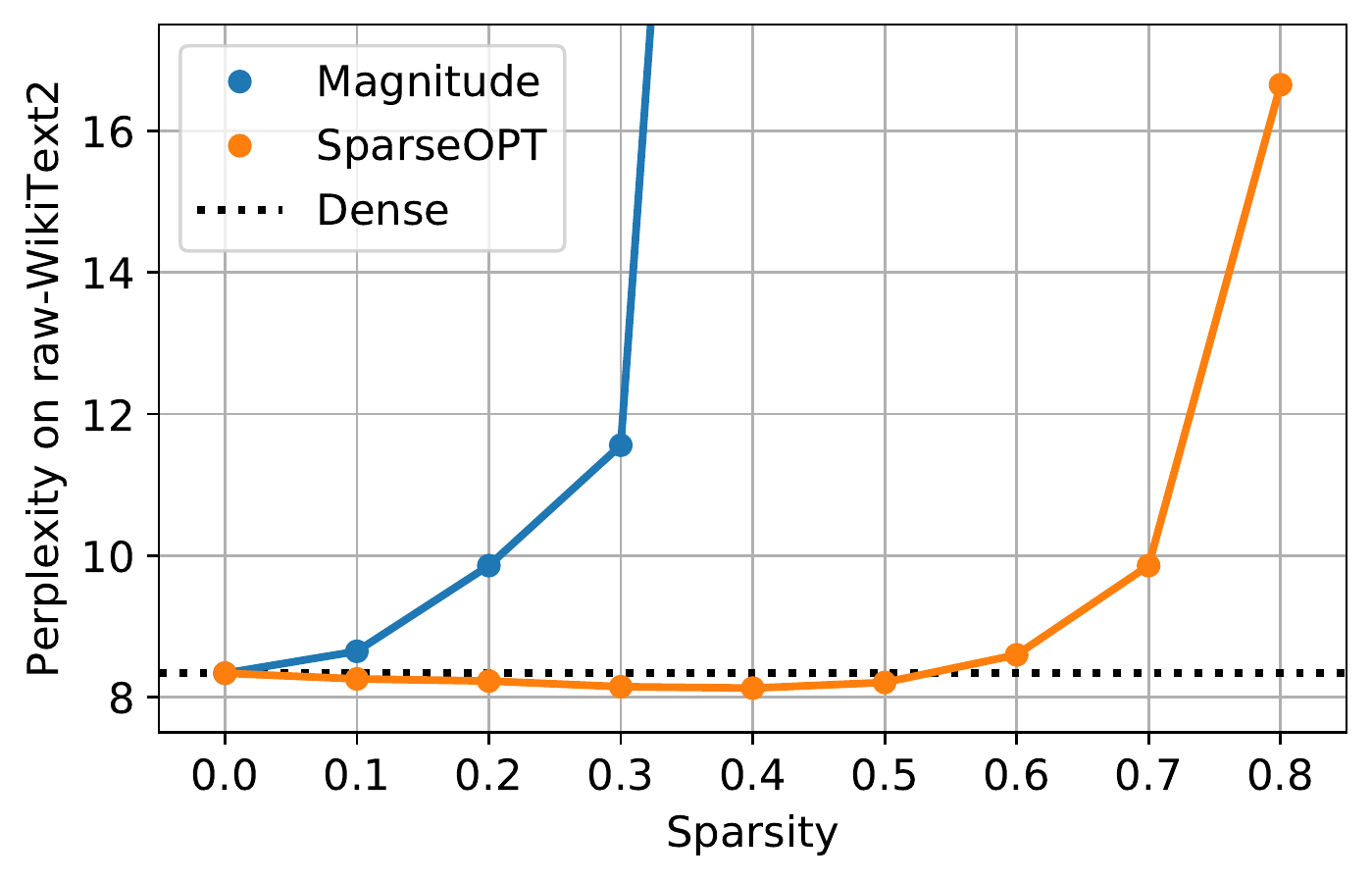}
        \vspace{-10pt}
        \captionof{figure}{Sparsity-vs-perplexity comparison of \sparsegpt{} against magnitude pruning on OPT-175B, when pruning to different \emph{uniform} per-layer sparsities.}
        \label{fig:opt-unstr}
    \end{minipage}
    \hfill
    \begin{minipage}[c]{.49\textwidth}
    \centering
        \includegraphics[width=.8\linewidth]{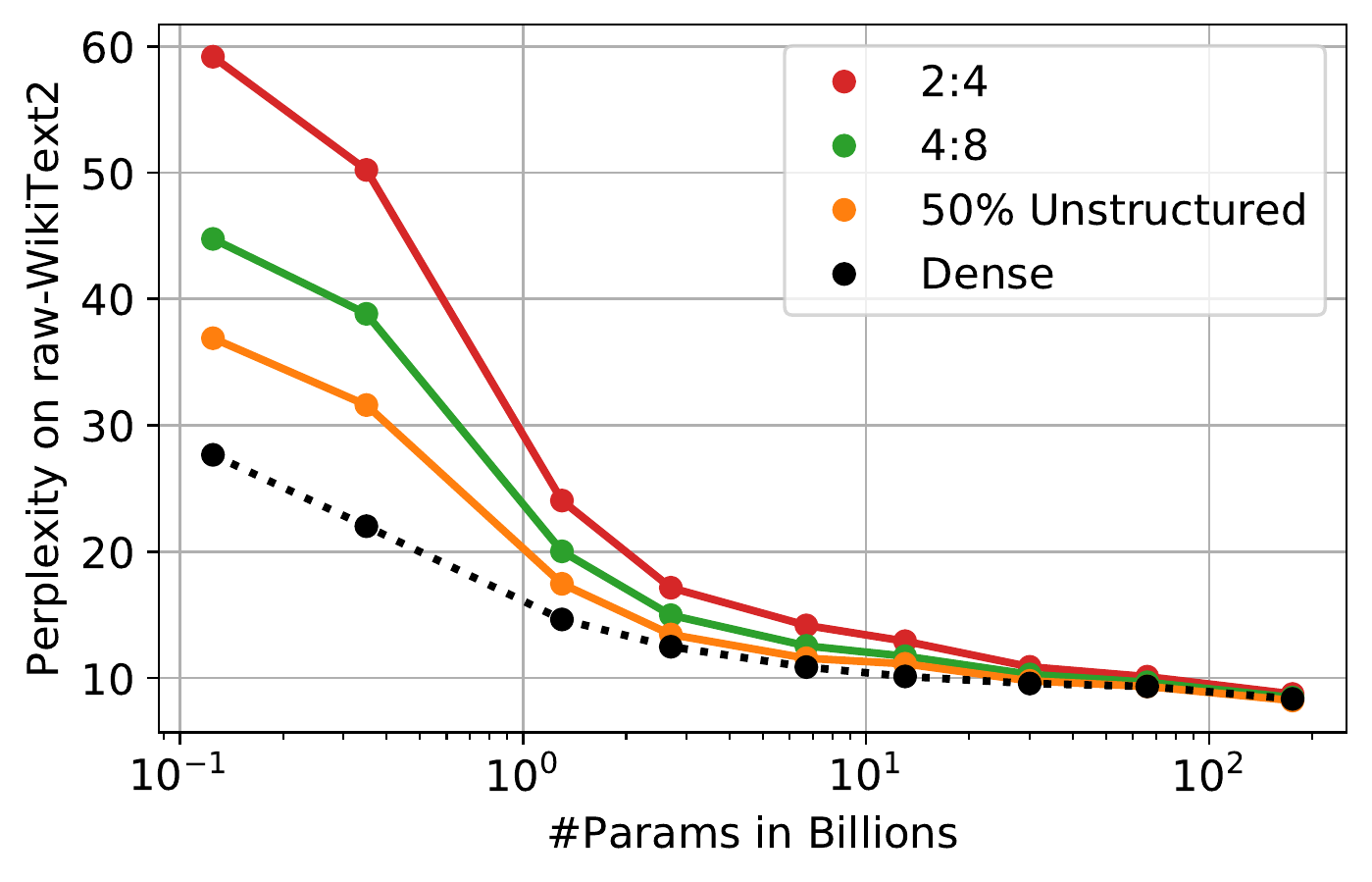}
        \vspace{-10pt}
        \captionof{figure}{Perplexity vs. model and sparsity type when compressing the entire OPT model family (135M, 350M, \ldots, 66B, 175B) to different sparsity patterns using \sparsegpt{}.}
        \label{fig:opt-sparsities}
    \end{minipage}
\end{figure*}

\textbf{Overview.} 
In this paper, we propose \sparsegpt{}, the first accurate one-shot pruning method which works efficiently at the scale of models with 10-100+ billion parameters. 
\sparsegpt{} works by reducing the pruning problem to a set of extremely large-scale instances of \emph{sparse regression}. It then solves these instances via a new approximate sparse regression solver, which is efficient enough to execute in a few hours on the largest openly-available GPT models (175B parameters), 
on a single GPU. 
At the same time, \sparsegpt{} is accurate enough to drop negligible accuracy post-pruning, without any fine-tuning.  
For example, when executed on the largest publicly-available generative language models (OPT-175B and BLOOM-176B), \sparsegpt{} induces 50-60\% sparsity in one-shot, with minor accuracy loss, measured either in terms of perplexity or zero-shot accuracy.  

Our experiments, from which we provide a snapshot in Figures \ref{fig:opt-unstr} and \ref{fig:opt-sparsities}, lead to the following observations. 
First, as shown in Figure~\ref{fig:opt-unstr}, \sparsegpt{} can induce uniform layer-wise sparsity of up to 60\% in e.g. the 175-billion-parameter variant of the OPT family~\cite{zhang2022opt}, with minor accuracy loss. 
By contrast, the only known one-shot baseline which easily extends to this scale, Magnitude Pruning~\cite{hagiwara1994, han2015learning}, preserves accuracy only until 10\% sparsity, and completely collapses beyond 30\% sparsity. 
Second, as shown in Figure~\ref{fig:opt-sparsities}, \sparsegpt{} can also accurately impose sparsity in the 
more stringent, but hardware-friendly, 2:4 and 4:8 semi-structured sparsity patterns~\cite{NVIDIASparse}, 
although this comes at an accuracy loss relative to the dense baseline for smaller models.  

One key positive finding, illustrated in Figure~\ref{fig:opt-sparsities}, is that \emph{larger models are more compressible}:  
 they drop significantly less accuracy at a fixed sparsity, relative to their smaller counterparts. 
 (For example, the largest models from the OPT and BLOOM families can be sparsified to 50\% with almost no increase in perplexity.) 
In addition, our method allows sparsity to be \emph{compounded} with weight quantization techniques~\citep{frantar2022gptq}: 
for instance, we can induce 50\% weight sparsity jointly with 4-bit weight quantization with negligible perplexity increase on OPT-175B. 

One notable property of \sparsegpt{} is that it is \emph{entirely local}, in the sense that it relies solely on weight updates designed to preserve the input-output relationship for each layer, which are computed without any global gradient information. As such, we find it remarkable that one can directly identify such sparse models in the ``neighborhood'' of dense pretrained models, whose output correlates extremely closely with that of the dense model.  

\section{Background}
\label{sec:preliminaries}

\paragraph{Post-Training Pruning} is a practical scenario where we are given a well-optimized model $\theta^{\star}$, together with some calibration data, and must obtain a compressed (e.g., sparse and/or quantized) version of $\theta^{\star}$. Originally popularized in the context of quantization~\cite{hubara2021accurate, nagel2020up, li2021brecq}, this setting has also recently been successfully extended to pruning~\cite{hubara2021accelerated, frantar2022obc, kwon2022fast}.

\paragraph{Layer-Wise Pruning.} Post-training compression is usually done by splitting the full-model compression problem into \emph{layer-wise} subproblems, whose solution quality is measured in terms of the $\ell_2$-error between the output, for given inputs $\mathbf{X_\ell}$, of the uncompressed layer with weights $\mathbf{W_\ell}$ and that of the compressed one. Specifically, for pruning, \cite{hubara2021accelerated} posed this problem as that of finding, for each layer $\ell$, a sparsity mask\footnote{Throughout the paper, by \emph{sparsity mask} for a given tensor we mean a binary tensor of the same dimensions, with $0$ at the indices of the sparsified entries, and $1$ at the other indices.} $\mathbf{M}_\ell$ with a certain target density, and possibly updated weights $\mathbf{\widehat{W}_\ell}$ such that 
\begin{equation}
    \label{eq:layerwise-pruning}
    \text{argmin}_{\textnormal{mask $\mathbf{M_\ell}$},\mathbf{\widehat{W}_\ell}} \, ||\mathbf{W_\ell} \mathbf{X_\ell} - (\mathbf{M_\ell} \odot  \mathbf{\widehat{W}_\ell}) \mathbf{X}_\ell||_2^2.
\end{equation}
The overall compressed model is then obtained by ``stitching together'' the individually compressed layers.

\paragraph{Mask Selection \& Weight Reconstruction.} A key aspect of the layer-wise pruning problem in (\ref{eq:layerwise-pruning}) is that both the mask $\mathbf{M}_\ell$ as well as the remaining weights $\widehat{\mathbf{W}}_\ell$ are optimized \emph{jointly}, which makes this problem NP-hard \cite{blumensath2008iterative}. Thus, exactly solving it for larger layers is unrealistic, leading all existing methods to resort to approximations.

A particularly popular approach is to separate the problem into \emph{mask selection} and \emph{weight reconstruction} \cite{he2018amc, kwon2022fast, hubara2021accelerated}. Concretely, this means to first choose a pruning mask $\mathbf{M}$ according to some saliency criterion, like the weight magnitude~\cite{zhu2017prune}, and then optimize the remaining unpruned weights while keeping the mask unchanged. Importantly, once the mask is fixed, (\ref{eq:layerwise-pruning}) turns into a \textit{linear squared error problem} that is easily optimized.

\paragraph{Existing Solvers.} Early work \cite{Kingdon1997} applied iterated linear regression to small networks. More recently, the AdaPrune approach \cite{hubara2021accelerated} has shown good results for this problem on modern models via magnitude-based weight selection, followed by applying SGD steps to reconstruct the remaining weights. Follow-up works demonstrate that pruning accuracy can be further improved by removing the strict separation between mask selection and weight reconstruction. Iterative AdaPrune \cite{frantar2022spdy} performs pruning in gradual steps with reoptimization in between and OBC \cite{frantar2022obc} introduces a greedy solver which removes weights one-at-a-time, fully reconstructing the remaining weights after each iteration, via efficient closed-form equations.

\paragraph{Difficulty of Scaling to 100+ Billion Parameters.} Prior post-training techniques have all been designed to accurately compress models up to a few hundred million parameters with several minutes to a few hours of compute. However, our goal here  is to sparsify models up to $1000\times$ larger.

Even AdaPrune, the method optimized for an ideal speed/accuracy trade-off, takes a few hours to sparsify models with just 1.3 billion parameters (see also Section~\ref{sec:experiments}), scaling linearly to several hundred hours (a few weeks) for 175B Transformers. More accurate approaches are at least several times more expensive \cite{frantar2022spdy} than AdaPrune or even exhibit worse than linear scaling \cite{frantar2022obc}. This suggests that scaling up existing accurate post-training techniques to extremely large models is a challenging endeavor. Hence, we propose a new layer-wise solver \sparsegpt{}, based on careful approximations to closed form equations, which easily scales to giant models, both in terms of runtime as well as accuracy.

\section{The \sparsegpt{} Algorithm}

\subsection{Fast Approximate Reconstruction}
\label{sec:fast-approximate-reconstruction}

\paragraph{Motivation.} As outlined in Section~\ref{sec:preliminaries}, for a fixed pruning mask $\mathbf{M}$, the optimal values of all weights in the mask can be calculated exactly by solving the sparse reconstruction problem corresponding to each matrix row $\mathbf{w^i}$ via:
\begin{equation}
    \label{eqn:rowwise-pruning}
    \mathbf{w^i}_{\mathbf{M_i}} = (\mathbf{X}_{\mathbf{M_i}} \mathbf{X}_{\mathbf{M_i}}^\top)^{-1} \mathbf{X}_{\mathbf{M_i}} (\mathbf{w}_{\mathbf{M_i}}\mathbf{X}_{\mathbf{M_i}})^\top,
\end{equation}
where $\mathbf{X}_{\mathbf{M_i}}$ denotes only the subset of input features whose corresponding weights have not been pruned in row $i$, and $\mathbf{w}_{\mathbf{M_i}}$ represents their respective weights. However, this requires inverting the Hessian matrix $ \mathbf{H}_{\mathbf{M_i}} = \mathbf{X}_{\mathbf{M_i}} \mathbf{X}_{\mathbf{M_i}}^\top$ corresponding to the values preserved by the pruning mask $\mathbf{M_i}$ for row $i$, i.e. computing $(\mathbf{H}_{\mathbf{M_i}})^{-1}$, separately for all rows $1 \leq i \leq d_\text{row}$. One such inversion takes $O(d_\text{col}^3)$ time, for a total computational complexity of $O(d_\text{row} \cdot d_\text{col}^3)$ over  $d_\text{row}$ rows. For a Transformer model, this means that the overall runtime scales with the 4th power of the hidden dimension $d_\text{hidden}$; we need a speedup by at least a full factor of $d_\text{hidden}$ to arrive at a practical algorithm.

\begin{figure}
  \begin{center}
    \includegraphics[width=0.85\linewidth]{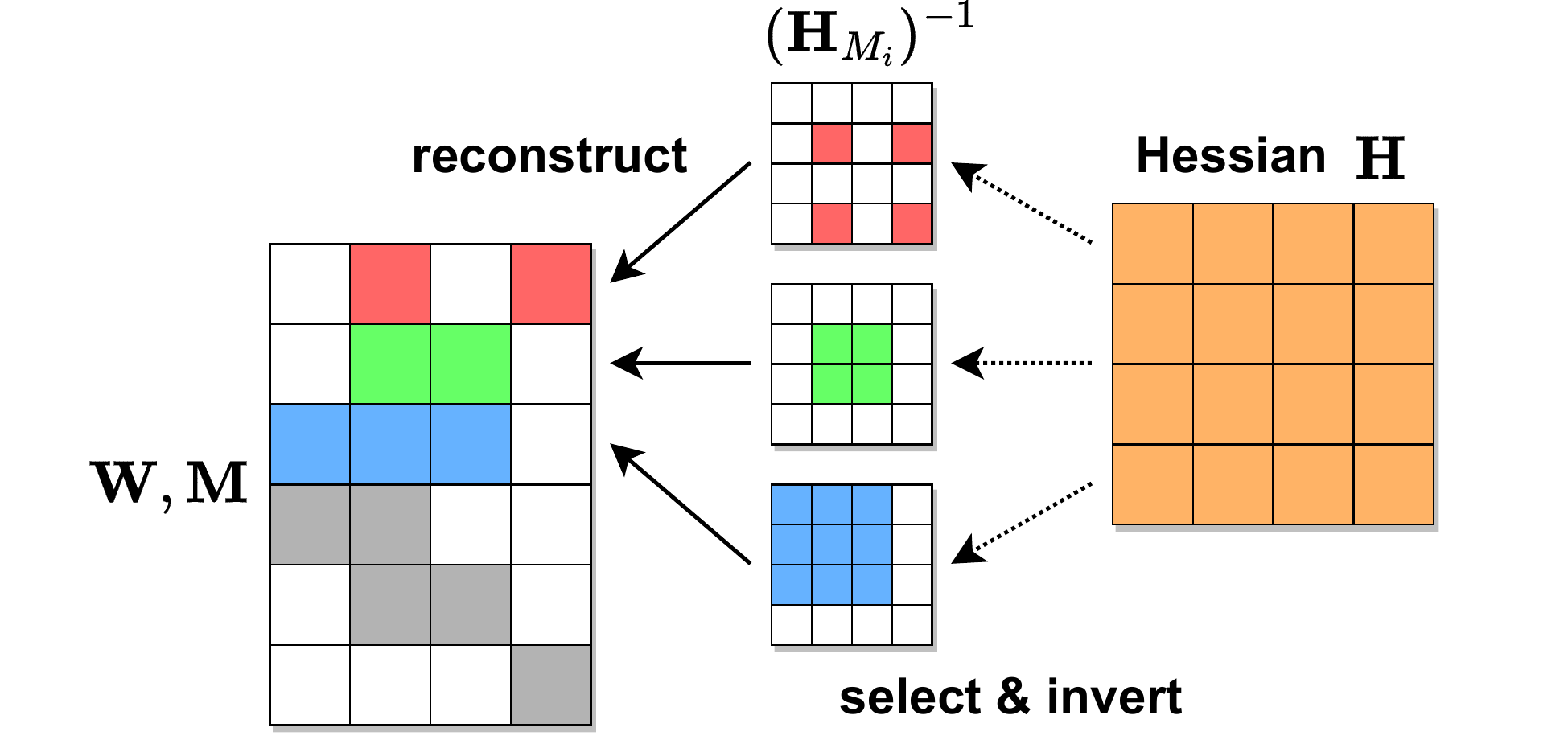}
  \end{center}
  \vspace{-10pt}
      \caption{Illustration of the row-Hessian challenge: rows are sparsified independently, pruned weights are in white.}
    \label{fig:different-row-hessians}
\end{figure}

\paragraph{Different Row-Hessian Challenge.} The high computational complexity of optimally reconstrucing the unpruned weights following Equation~\ref{eqn:rowwise-pruning} mainly stems from the fact that solving \emph{each row} requires the \emph{individual} inversion of a $O(d_\text{col} \times d_\text{col})$ matrix. This is because the row masks $\mathbf{M_i}$ are generally different and $(\mathbf{H}_{\mathbf{M_i}})^{-1} \neq (\mathbf{H}^{-1})_{\mathbf{M_i}}$, i.e., the inverse of a masked Hessian does \emph{not} equal the masked version of the full inverse. This is illustrated also in Figure~\ref{fig:different-row-hessians}. If all row-masks were the same, then we would only need to compute a single shared inverse, as $\mathbf{H} = \mathbf{X}\mathbf{X}^\top$ depends just on the layer inputs which are the same for all rows. 

Such a constraint could be enforced in the mask selection, but this would have a major impact on the final model accuracy, as sparsifying weights in big structures, like entire columns, is known to be much more difficult than pruning them individually\footnote{For example, structured (column-wise) pruning ResNet50 to $> 50$\% structured sparsity without accuracy loss is challenging, even with extensive retraining \cite{liu2021group}, while  unstructured pruning to 90\% sparsity is easily achievable with state-of-the-art methods~\cite{evci2020rigging, peste2021ac}.}. The key towards designing an approximation algorithm that is both accurate and efficient lies in enabling the reuse of Hessians between rows with distinct pruning masks. We now propose an algorithm that achieves this in a principled manner.

\begin{figure*}[h]
    \centering
    \includegraphics[width=.8\textwidth]{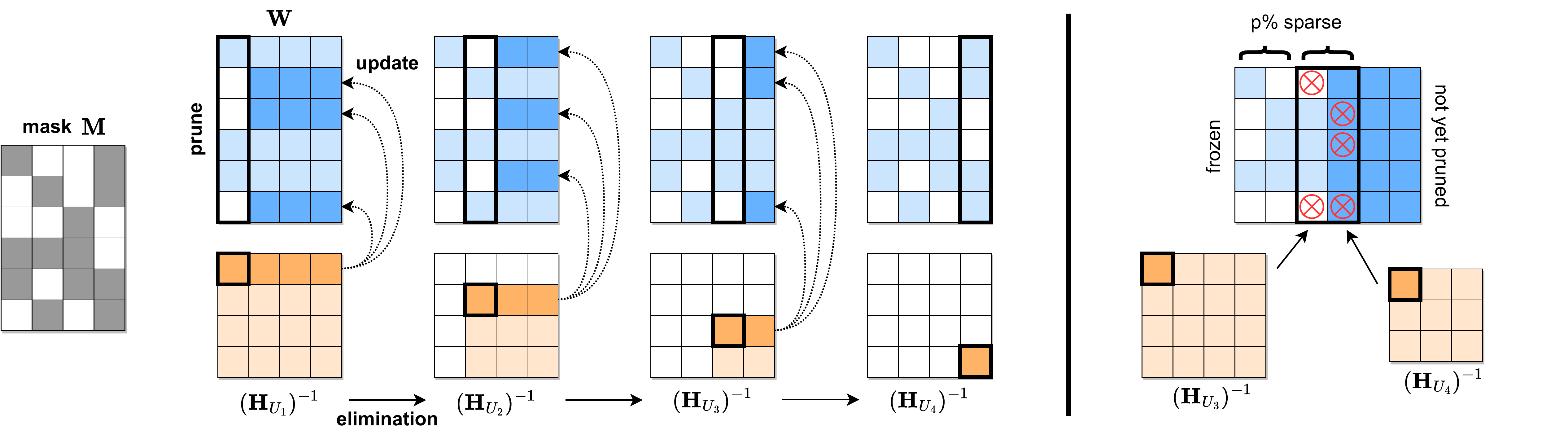}
    \vspace{-5pt}
    \caption{[Left] Visualization of the \sparsegpt{} reconstruction algorithm. Given a fixed pruning mask $\mathbf{M}$, we incrementally prune weights in each column of the weight matrix $\mathbf{W}$, using a sequence of Hessian inverses $(\mathbf{H}_{U_j})^{-1}$, and updating the remainder of the weights in those rows, located to the ``right'' of the column being processed. Specifically, the weights to the ``right'' of a pruned weight (dark blue) will be updated to compensate for the pruning error, whereas the unpruned weights do not generate updates (light blue). [Right] Illustration of the adaptive mask selection via iterative blocking.}
    \label{fig:sparse-gpt-vis}
\end{figure*}

\paragraph{Equivalent Iterative Perspective.} To motivate our algorithm, we first have to look at the row-wise weight reconstruction from a different \emph{iterative} perspective, using the classic OBS update \cite{hassibi1993optimal, singh2020woodfisher, frantar2021m}. Assuming a quadratic approximation of the loss, for which the current weights $\mathbf{w}$ are optimal, the OBS update $\boldsymbol{\delta}_m$ provides the optimal adjustment of the remaining weights to compensate for the removal of the weight at index $m$, incurring error $\varepsilon_m$:
\begin{equation}
    \label{eq:obs-update}
    \boldsymbol{\delta_m} = - \frac{w_m}{[\mathbf{H}^{-1}]_{mm}} \cdot \mathbf{H}^{-1}_{:, m}, \quad \varepsilon_m = \frac{w_m^2}{[\mathbf{H}^{-1}]_{mm}}.
\end{equation}
Since the loss function corresponding to the layer-wise pruning of one row of $\mathbf{W}$ is a quadratic, the OBS formula is exact in this case. Hence, $\mathbf{w} + \boldsymbol{\delta}_m$ is the optimal weight reconstruction corresponding to mask $\{m\}^C$. Further, given an optimal sparse reconstruction $\mathbf{w}^{(\mathbf{M})}$ corresponding to mask $\mathbf{M}$, we can apply OBS again to find the optimal reconstruction for mask $\mathbf{M}' = \mathbf{M} - \{m\}$. Consequently, this means that instead of solving for a full mask $\mathbf{M} = \{m_1, \dots, m_p\}^C$ directly, we could iteratively apply OBS to individually prune the weights $m_1$ up until $m_p$ in order, one-at-a-time, reducing an initially complete mask to $\mathbf{M}$, and will ultimately arrive at \emph{the same} optimal solution as applying the closed-form regression reconstruction with the full $\mathbf{M}$ directly. 

\textbf{Optimal Partial Updates.} Applying the OBS update $\boldsymbol{\delta}_m$ potentially adjusts the values of all available parameters (in the current mask $\mathbf{M}$) in order to compensate for the removal of $w_m$. However, what if we only update the weights in a subset $\mathbf{U} \subseteq \mathbf{M}$ among remaining unpruned weights? Thus, we could still benefit from error compensation, using only weights in $\mathbf{U}$, while reducing the cost of applying OBS.

Such a partial update can indeed be accomplished by simply computing the OBS update using $\mathbf{H}_{\mathbf{U}}$, the Hessian corresponding to $\mathbf{U}$, rather than $\mathbf{H}_{\mathbf{M}}$, and updating only $\mathbf{w}_{\mathbf{U}}$. Importantly, the loss of our particular layer-wise problem remains quadratic also for $\mathbf{U}$ and the OBS updates are still optimal: the restriction to $\mathbf{U}$ does not incur any extra approximation error by itself, only the error compensation might not be as effective, as less weights are available for adjustment. At the same time, if $|\mathbf{U}| < |{\mathbf{M}}|$, then inverting $\mathbf{H}_{{\mathbf{U}}}$ will be a lot faster than inverting $\mathbf{H}_{\mathbf{M}}$. We will now utilize this mechanism to accomplish our goal of synchronizing the masked Hessians across all rows of $\mathbf{W}$.

\textbf{Hessian Synchronization.} In the following, assume a fixed ordering of the input features $j = 1, \dots, d_\text{col}$. Since those are typically arranged randomly, we will just preserve the given order for simplicity, but any permutation could in principle be chosen. Next, we define a sequence of $d_\text{col}$ index subsets $U_j$ recursively as:
\begin{equation}
    U_{j + 1} = U_j - \{j\} \,\, \text{with} \,\, U_1 = \{1, \dots, d_\text{col}\}.
\end{equation}
In words, starting with $U_1$ being the set of all indices, each subset $U_{j + 1}$ is created by removing the smallest index from the previous subset $U_j$. These subsets also impose a sequence of inverse Hessians $(\mathbf{H}_{U_j})^{-1} = ((\mathbf{X}\mathbf{X}^\top)_{U_j})^{-1}$ which we are going to share across all rows of $\mathbf{W}$. Crucially, following \cite{frantar2022obc}, the updated inverse $(\mathbf{H}_{U_{j + 1}})^{-1}$ can be calculated efficiently by removing the first row and column, corresponding to $j$ in the original $\mathbf{H}$, from $\mathbf{B} = (\mathbf{H}_{U_j})^{-1}$ in $O(d_\text{col}^2)$ time via one step of Gaussian elimination:
\begin{equation}
    (\mathbf{H}_{U_{j + 1}})^{-1} = \Big(\mathbf{B} - \frac{1}{[\mathbf{B}]_{11}} \cdot \mathbf{B}_{:,1}\mathbf{B}_{1,:} \Big)_{2:,2:},
\end{equation}
\noindent with $(\mathbf{H}_{U_1})^{-1} = \mathbf{H}^{-1}.$
Hence, the entire sequence of $d_\text{col}$ inverse Hessians can be calculated recursively in $O(d_\text{col}^3)$ time, i.e. at similar cost to a single extra matrix inversion on top of the initial one for $\mathbf{H}^{-1}$.

Once some weight $w_k$ has been pruned, it should not be updated anymore. Further, when we prune $w_k$, we want to update as many unpruned weights as possible for maximum error compensation. This leads to the following strategy: iterate through the $U_j$ and their corresponding inverse Hessians $(\mathbf{H}_{U_j})^{-1}$ in order and prune $w_j$ if $j \not \in M_i$, for all rows $i$. Importantly, each inverse Hessian $(\mathbf{H}_{U_j})^{-1}$ is computed only once and reused to remove weight $j$ in all rows where it is part of the pruning mask. A visualization of the algorithm can be found in Figure \ref{fig:sparse-gpt-vis}.

\textbf{Computational Complexity.} 
The overall cost consists of three parts: (a) the computation of the initial Hessian, which takes time $\Theta(n \cdot d_\text{col}^2)$ where $n$ is the number of input samples used---we found that taking the number of samples $n$ to be a small multiple of $d_\text{col}$ is sufficient for good and stable results, even on very large models (see Appendix \ref{app:ablations}); (b) iterating through the inverse Hessian sequence in time $O(d_\text{col}^3)$ and (c) the reconstruction/pruning itself. The latter cost can be upper bounded by the time it takes to apply~(\ref{eq:obs-update}) to all $d_\text{row}$ rows of $\mathbf{W}$ for all $d_\text{col}$ columns in turn, which is $O(d_\text{col} d_\text{row} d_\text{col})$. In total, this sums up to $O(d_\text{col}^3 + d_\text{row} d_\text{col}^2)$. For Transformer models, this is simply $O(d_\text{hidden}^3)$, and is thus a full $d_\text{hidden}$-factor more efficient than exact reconstruction. This means that we have reached our initial goal, as this complexity will be sufficient to make our scheme practical, even for extremely large models.

\textbf{Weight Freezing Interpretation.} While we have motivated the \sparsegpt{} algorithm as an  approximation to the exact reconstruction using optimal partial updates, there is also another interesting view of this scheme. Specifically, consider an exact greedy framework which compresses a weight matrix column by column, always optimally updating all not yet compressed weights in each step \cite{frantar2022obc, frantar2022gptq}. At first glance, \sparsegpt{} does not seem to fit into this framework as we only compress some of the weights in each column and also only update a subset of the uncompressed weights. Yet, mechanically, ``compressing'' a weight ultimately means fixing it to some specific value and ensuring that it is never ``decompressed'' again via some future update, i.e. that it is \emph{frozen}. Hence, by defining column-wise compression as:
\begin{equation}
\text{compress}(\mathbf{w^j})_i = 0 \,\, \text{if} \,\, j \not \in M_i \,\, \text{and} \,\, w^j_i \,\, \text{otherwise},
\end{equation}
i.e. zeroing weights not in the mask and fixing the rest to their current value, our algorithm can be interpreted as an exact column-wise greedy scheme. This perspective will allow us to cleanly merge sparsification and quantization into a single compression pass.

\subsection{Adaptive Mask Selection}

So far, we have focused only on weight reconstruction, i.e. assuming a fixed pruning mask $\mathbf{M}$. One simple option for deciding the mask, following AdaPrune \cite{hubara2021accelerated}, would be via magnitude pruning~\cite{zhu2017prune}. 
However, recent work \cite{frantar2022obc} shows that updates during pruning change weights significantly due to correlations, and that taking this into account in the mask selection yields better results. 
This insight can be integrated into \sparsegpt{} by \emph{adaptively} choosing the mask while running the reconstruction.

One obvious way of doing so would be picking the $p\%$ easiest weights to prune in each column $i$ when it is compressed, leading to $p\%$ overall sparsity. 
The big disadvantage of this approach is that sparsity cannot be distributed non-uniformly across columns, imposing additional unnecessary structure. 
This is particularly problematic for massive language models, which have a small number of highly-sensitive outlier features~\cite{dettmers2022llm, xiao2022smoothquant}.

We remove this disadvantage via \textit{iterative blocking}. 
More precisely, we always select the pruning mask for $B_s = 128$ columns at a time (see Appendix \ref{app:ablations}), based on the OBS reconstruction error $\varepsilon$ from Equation~(\ref{eq:obs-update}), using the diagonal values in our Hessian sequence. 
We then perform the next $B_s$ weight updates, before selecting the mask for the next block, and so on. 
This procedure allows \emph{non-uniform selection} per column, in particular also using the corresponding Hessian information, while at the same time considering also previous weight updates for selection.
(For a single column $j$, the selection criterion becomes the magnitude, as $[\mathbf{H}^{-1}]_{jj}$ is constant across rows.)

\subsection{Extension to Semi-Structured Sparsity} 

\sparsegpt{} is also easily adapted to \textit{semi-structured} patterns such as the popular n:m sparsity format \cite{zhou2021learning, hubara2021accelerated} which delivers speedups in its 2:4 implementation on Ampere NVIDIA GPUs. Specifically, every consecutive $m$ weights should contain exactly $n$ zeros. Hence, we can simply choose blocksize $B_s = m$ and then enforce the zeros-constraint in the mask selection for each row by picking the $n$ weights which incur the lowest error as per Equation (\ref{eq:obs-update}). A similar strategy could also be applied for other semi-structured pruning patterns. Finally, we note that a larger $B_s$ would not be useful in this semi-structured scenario since zeros cannot be distributed non-uniformly between different column-sets of size $m$.

\subsection{Full Algorithm Pseudocode}

\begin{algorithm}[h!]
    \centering
    \caption{The \sparsegpt{} algorithm. We prune the layer matrix $\mathbf{W}$ to $p\%$ unstructured sparsity given inverse Hessian $\mathbf{H}^{-1} = (\mathbf{X} \mathbf{X}^\top + \lambda \mathbf{I})^{-1}$, lazy batch-update blocksize $B$ and adaptive mask selection blocksize $B_s$; each $B_s$ consecutive columns will be $p\%$ sparse.}
    \small
    \label{alg:sparsegpt}
    \begin{algorithmic}
        \STATE $\mathbf{M} \gets \mathbf{1}_{d_\text{row} \times d_\text{col}}$ \,\, \textit{// binary pruning mask}
        \STATE $\mathbf{E} \gets \mathbf{0}_{d_\text{row} \times B}$ \,\, \textit{// block quantization errors}
        \STATE $\mathbf{H}^{-1} \gets \text{Cholesky}
        (\mathbf{H}^{-1})^\top$ \,\, \textit{// Hessian inverse information}
        \FOR {$i = 0, B, 2B, \dots$}
            \FOR {$j = i, \dots, i + B - 1$}
                {
                \IF {$j \, \text{mod} \, B_s = 0$}
                    \STATE $\mathbf{M}_{:, j:(j+B_s)} \gets$ mask of $(1 - p)\%$ weights $w_c \in \mathbf{W}_{:, j:(j + B_s)}$ with largest $w_c^2 / [\mathbf{H}^{-1}]_{cc}^2$
                \ENDIF
                \STATE $\mathbf{E}_{:, j - i} \gets \mathbf{W}_{:, j} \, / \, [\mathbf{H}^{-1}]_{jj}$ \,\, \textit{// pruning error}
                \STATE $\mathbf{E}_{:, j - i} \gets (\mathbf{1} - \mathbf{M}_{:,j}) \cdot \mathbf{E}_{:, j - i}$ \,\, \textit{// freeze weights}
                }
                \STATE $\mathbf{W}_{:, j:(i + B)} \gets \mathbf{W}_{:, j:(i + B)} - \mathbf{E}_{:, j - i} \cdot \mathbf{H}^{-1}_{j, j:(i + B)}$ \,\, \textit{// update}
            \ENDFOR
            \STATE $\mathbf{W}_{:, (i + B):} \gets \mathbf{W}_{:, (i + B):} - \mathbf{E} \cdot \mathbf{H}^{-1}_{i:(i + B), (i + B):}$ \,\, \textit{// update}
        \ENDFOR
        \STATE {$\mathbf{W} \gets \mathbf{W} \cdot \mathbf{M}$ \,\, \textit{// set pruned weights to 0}}
    \end{algorithmic}
\end{algorithm}

\begin{table*}[h!]
    \centering
    \caption{OPT perplexity results on raw-WikiText2.}
    \vspace{5pt}
    \scalebox{.85}{
        \begin{tabular}{|l|c|c|c|}
            \toprule
            OPT - 50\% & 125M & 350M & 1.3B \\
            \midrule
            Dense & 27.66 & 22.00 & 14.62 \\
            \midrule
            Magnitude & 193. & 97.80 & 1.7e4 \\
            AdaPrune & 58.66 & 48.46 & 32.52 \\
            \sparsegpt{} &\textbf{36.85} & \textbf{31.58} & \textbf{17.46} \\
            \bottomrule
        \end{tabular}
    }~~~
    \scalebox{.8}{
        \begin{tabular}{|l|c|c|c|c|c|c|c|}
            \toprule
            OPT & Sparsity & 2.7B & 6.7B & 13B & 30B & 66B & 175B \\
            \midrule
            Dense & 0\% & 12.47 & 10.86 & 10.13 & 9.56 & 9.34 & 8.35 \\
            \midrule
            Magnitude & 50\% & 265. & 969. & 1.2e4 & 168. & 4.2e3 & 4.3e4 \\
            \sparsegpt{} & 50\% & \textbf{13.48} & \textbf{11.55} & \textbf{11.17} & \textbf{9.79} & \textbf{9.32} & \textbf{8.21} \\
            \midrule
            \sparsegpt{} & 4:8 & 14.98 & 12.56 & 11.77 & 10.30 & 9.65 & 8.45 \\
            \sparsegpt{} & 2:4 & 17.18 & 14.20 & 12.96 & 10.90 & 10.09 & 8.74 \\
            \bottomrule
        \end{tabular}
    }
    \label{tab:opt-wikitext2}
\end{table*}

With the weight freezing interpretation discussed at the end of Section \ref{sec:fast-approximate-reconstruction}, the \sparsegpt{} reconstruction can be cast in the column-wise greedy framework of the recent quantization algorithm GPTQ \cite{frantar2022gptq}. This means we can also inherit several algorithmic enhancements from GPTQ, specifically: precomputing all the relevant inverse Hessian sequence information via a Cholesky decomposition to achieve numerical robustness and applying lazy batched weight matrix updates to improve the compute-to-memory ratio of the algorithm. Our adaptive mask selection, as well as its extensions to semi-structured pruning, are compatible with all of those extra techniques as well.

Algorithm~\ref{alg:sparsegpt} presents the the unstructured sparsity version of the \sparsegpt{} algorithm in its fully-developed form, integrating all the relevant techniques from GPTQ. 

\subsection{Joint Sparsification \& Quantization}
\label{sec:joint}

Algorithm \ref{alg:sparsegpt} operates in the column-wise greedy framework of GPTQ, thus sharing the computationally heavy steps of computing the Cholesky decomposition of $\mathbf{H}^{-1}$ and continuously updating $\mathbf{W}$. This makes it possible to merge both algorithms into a single joint procedure. Specifically, all weights that are frozen by \sparsegpt{} are additionally quantized, leading to the following generalized errors to be compensated in the subsequent update step:
\begin{equation}
    \mathbf{E}_{:, j - i} \gets (\mathbf{W}_{:, j} - \mathbf{M}_{:, j} \cdot \text{quant}(\mathbf{W}_{:, j})) \, / \, [\mathbf{H}^{-1}]_{jj},
\end{equation}
where $\text{quant}(\mathbf{w})$ rounds each weight in $\mathbf{w}$ to the nearest value on the quantization grid. Crucially, in this scheme, sparsification and pruning are performed \textit{jointly} in \textit{a single pass} at essentially no extra cost over \sparsegpt{}. Moreover,  doing quantization and pruning jointly means that later pruning decisions are influenced by earlier quantization rounding, and vice-versa. This is in contrast to prior joint techniques~\cite{frantar2022obc}, which first sparsify a layer and then simply quantize the remaining weights.

\section{Experiments}
\label{sec:experiments}

\paragraph{Setup.} We implement \sparsegpt{} in PyTorch \cite{paszke2019pytorch} and use the HuggingFace Transformers library~\cite{wolf2019huggingface} for handling models and datasets. All pruning experiments are conducted on \emph{a single} NVIDIA A100 GPU with 80GB of memory. In this setup, \sparsegpt{} can fully sparsify the 175-billion-parameter models in approximately 4 hours.  Similar to \citet{yao2022zeroquant, frantar2022gptq}, we sparsify Transformer layers sequentially in order, which significantly reduces memory requirements. All our experiments are performed in one-shot, without finetuning, in a similar setup to recent work on post-training quantization of GPT-scale models~\cite{frantar2022gptq, yao2022zeroquant, dettmers2022llm}. Additionally, in Appendix \ref{app:speedup} we investigate the real-world acceleration of our sparse models with existing tools.

For calibration data, we follow~\citet{frantar2022gptq} and use 128 2048-token segments, randomly chosen from the first shard of the C4 \cite{C4} dataset. 
This represents generic text data crawled from the internet and makes sure that our experiments remain actually zero-shot since \emph{no task-specific data is seen during pruning}.

\paragraph{Models, Datasets \& Evaluation.} We primarily work with the OPT model family \cite{zhang2022opt}, to study scaling behavior, but also consider the 176 billion parameter version of BLOOM \cite{scao2022bloom}. While \emph{our focus lies on the very largest variants}, we also show some results on smaller models to provide a broader picture.

In terms of metrics, we mainly focus on \textit{perplexity}, which is known to be a challenging and stable metric that is well suited for evaluating the accuracy of compression methods \cite{yao2022zeroquant, frantar2022obc, dettmers2022case}. We consider the test sets of raw-WikiText2 \cite{wikitext103} and PTB \cite{PTB} as well as a subset of the C4 validation data, all popular benchmarks in LLM compression literature \cite{yao2022zeroquant, park2022nuqmm, frantar2022gptq, xiao2022smoothquant}. For additional interpretability, we also provide ZeroShot accuracy results for Lambada~\cite{paperno2016lambada}, ARC (Easy and Challenge)~\cite{boratko2018systematic},  PIQA~\cite{tata2003piqa} and StoryCloze~\cite{mostafazadeh2017lsdsem}.

We note that the main focus of our evaluation lies on the \emph{accuracy of the sparse models, relative to the dense baseline} rather than on absolute numbers. Different preprocessing may influence absolute accuracy, but has little impact on our relative claims.  The perplexity is calculated following precisely the procedure described by HuggingFace~\cite{hfperplexity}, using full stride. Our ZeroShot evaluations are performed with GPTQ's \cite{frantar2022gptq} implementation, which is in turn based on the popular EleutherAI-eval-harness~\cite{eleuther}. Additional evaluation details can be found in Appendix \ref{app:evaluation-details}. All dense and sparse results were computed with exactly the same code, available as supplementary material, to ensure a fair comparison.

\paragraph{Baselines.} We compare against the standard magnitude pruning baseline \cite{zhu2017prune}, applied layer-wise, which scales to the very largest models. 
On models up to 1B parameters, we compare also against AdaPrune \cite{hubara2021accelerated}, the most efficient among existing accurate post-training pruning methods. 
For this, we use the memory-optimized reimplementation of~\citet{frantar2022spdy} and further tune the hyper-parameters provided by the AdaPrune authors. We thus achieve a $\approx3\times$ speedup without impact on solution quality, for our models of interest.
\vspace{-15pt}

\subsection{Results}
\label{sec:results}

\paragraph{Pruning vs. Model Size.} We first study how the difficulty of pruning LLMs changes with their size. We consider the entire OPT model family and uniformly prune all linear layers, excluding the embeddings and the head, as standard \cite{2020-sanh, kurtic2022optimal}, to 50\% unstructured sparsity, full 4:8 or full 2:4 semi-structured sparsity (the 2:4 pattern is the most stringent). The raw-WikiText2 performance numbers are given in Table \ref{tab:opt-wikitext2} and visualized in Figure \ref{fig:opt-sparsities}. The corresponding results for PTB and C4 can be found in Appendix \ref{app:additional-experiments} and show very similar trends overall.

\begin{figure*}[ht]
    \begin{minipage}[c]{.325\textwidth}
        \centering
        \includegraphics[width=.95\linewidth]{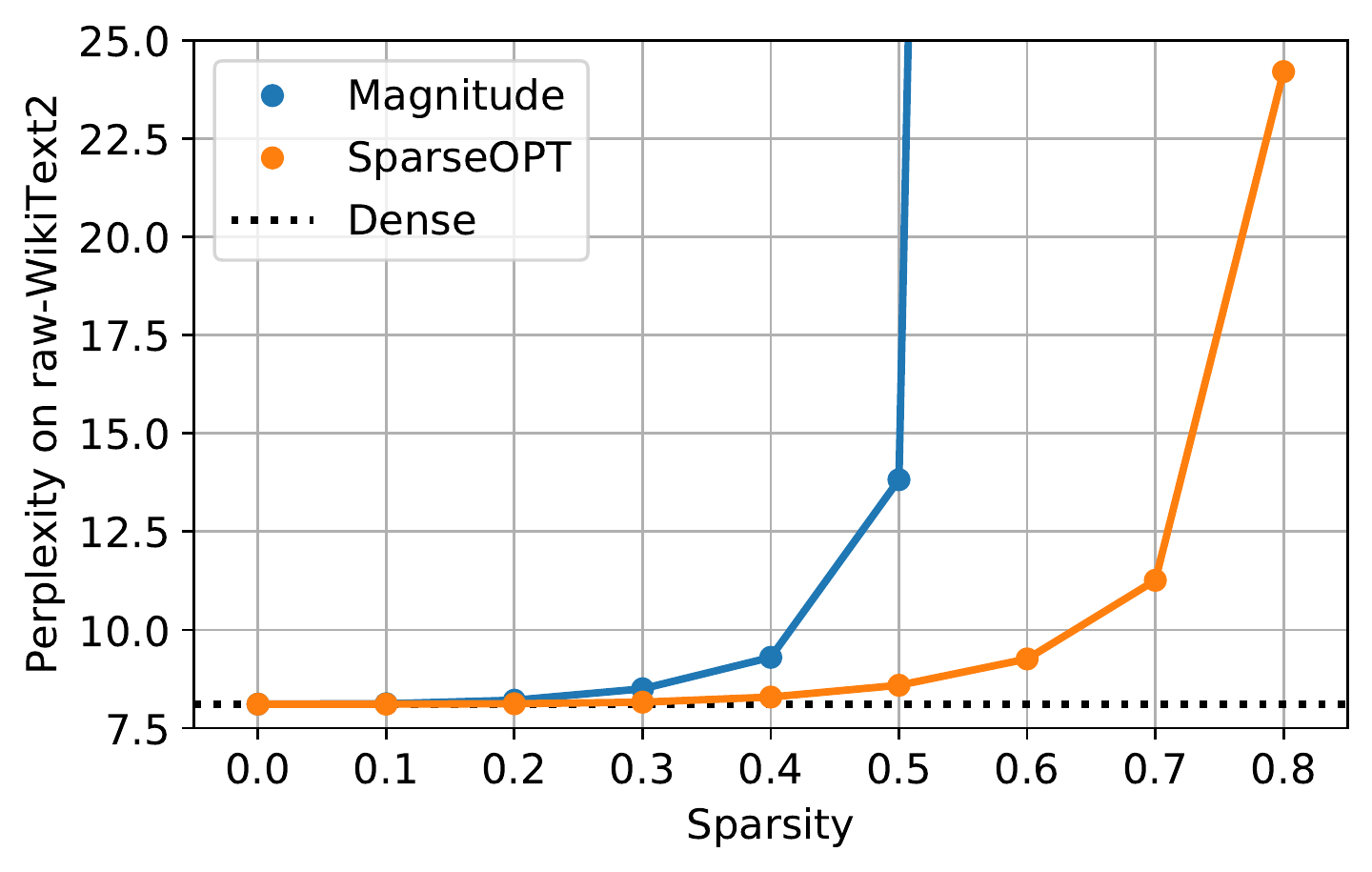}
        \vspace{-10pt}
        \captionof{figure}{Uniform pruning BLOOM-176B.}
        \label{fig:bloom-unstr}
    \end{minipage}
    \hfill
    \begin{minipage}[c]{.675\textwidth}
        \centering
        \vspace{-15pt}
        \captionof{table}{ZeroShot results on several datasets for sparsified variants of OPT-175B.}
        \vspace{5pt}
        \scalebox{.85}{
            \begin{tabular}{|l|c|c|c|c|c|c|c|}
                \toprule
                Method & Spars. & Lamb. & PIQA & ARC-e & ARC-c & Story. & \textbf{Avg.} \\
                \midrule
                Dense & 0\% & 75.59 & 81.07 & 71.04 & 43.94 & 79.82 & \textbf{70.29} \\
                \midrule
                Magnitude & 50\% & 00.02 & 54.73 & 28.03 & 25.60 & 47.10 & \textbf{31.10} \\
                \midrule
                \sparsegpt{} & 50\% & 78.47 & 80.63 & 70.45 & 43.94 & 79.12 & \textbf{70.52} \\
                \sparsegpt{} & 4:8 & 80.30 & 79.54 & 68.85 & 41.30 & 78.10 & \textbf{69.62} \\
                \sparsegpt{} & 2:4 & 80.92 & 79.54 & 68.77 & 39.25 & 77.08 & \textbf{69.11} \\
                \bottomrule
            \end{tabular}
        }
        \label{tab:zeroshot}
    \end{minipage}
\end{figure*}

One immediate finding is that the accuracy of magnitude-pruned models collapses across all scales, with larger variants generally dropping faster than smaller ones. This is in stark contrast to smaller vision models which can usually be pruned via simple magnitude selection to 50\% sparsity or more at very little loss of accuracy~\cite{singh2020woodfisher, frantar2022obc}. It highlights the importance of accurate pruners for massive generative language models, but also the fact that perplexity is a very sensitive metric. 

For \sparsegpt{}, the trend is very different: already at 2.7B parameters, the perplexity loss is $\approx 1$ point, at 66B, there is essentially zero loss and at the very largest scale there is even a slight accuracy improvement over the dense baseline, which however seems to be dataset specific (see also Appendix \ref{app:additional-experiments}). AdaPrune, as expected, also yields a big improvement over magnitude pruning, but is significantly less accurate than \sparsegpt{}. Despite the efficiency of AdaPrune, running it takes approximately $\approx 1.3$h on a 350M model and $\approx 4.3$h on a 1.3B one, while \sparsegpt{} can fully sparsify 66B and 175B models in roughly the same time, executing on the same A100 GPU.

In general, there is a clear trend of larger models being easier to sparsify, which we speculate is due to overparametrization. 
A detailed investigation of this phenomenon would be a good direction for future work. 
For 4:8 and 2:4 sparsity, the behavior is similar, but accuracy drops are typically higher due to the sparsity patterns being more constrained \cite{hubara2021accelerated}. Nevertheless, at the largest scale, the perplexity increases are only of $0.11$ and $0.39$ for 4:8 and 2:4 sparsity, respectively.

\paragraph{Sparsity Scaling for 100+ Billion Parameter Models.} Next, we take a closer look at the largest publicly-available dense models, OPT-175B and BLOOM-176B, and investigate how their performance scales with the degree of sparsity induced by either \sparsegpt{} or magnitude pruning. The results are visualized in Figures \ref{fig:opt-unstr} and \ref{fig:bloom-unstr}.

For the OPT-175B model (Figure~\ref{fig:opt-unstr}) magnitude pruning can achieve at most 10\% sparsity before significant accuracy loss occurs; meanwhile, \sparsegpt{} enables up to 60\% sparsity at a comparable perplexity increase. BLOOM-176B (Figure~\ref{fig:bloom-unstr}) appears to be more favorable for magnitude pruning, admitting up 30\% sparsity without major loss; still, \sparsegpt{} can deliver 50\% sparsity, a $1.66\times$ improvement, at a similar level of perplexity degradation. Even at 80\% sparsity, models compressed by \sparsegpt{} still score reasonable perplexities, while magnitude pruning leads to a complete collapse ($>$100 perplexity) already at 40/60\% sparsity for OPT and BLOOM, respectively. Remarkably, \sparsegpt{} removes around \emph{100 billion weights} from these models, with low impact on accuracy.

\paragraph{ZeroShot Experiments.} To complement the perplexity evaluations, we provide results on several ZeroShot tasks. These evaluations are known to be relatively noisy \cite{dettmers2022llm}, but more interpretable. Please see Table~\ref{tab:zeroshot}.

Overall, a similar trend holds, with magnitude-pruned models collapsing to close to random performance, while \sparsegpt{} models stay close to the original accuracy. However, as expected, these numbers are more noisy: 2:4 pruning appears to achieve noticeably higher accuracy than the dense model on Lambada, despite being the most constrained sparsity pattern. These effects ultimately average out when considering many different tasks, which is consistent to the literature~\cite{yao2022zeroquant, dettmers2022llm, dettmers2022case}.

\begin{figure}[h!]
    \centering
    \includegraphics[width=.8\linewidth]{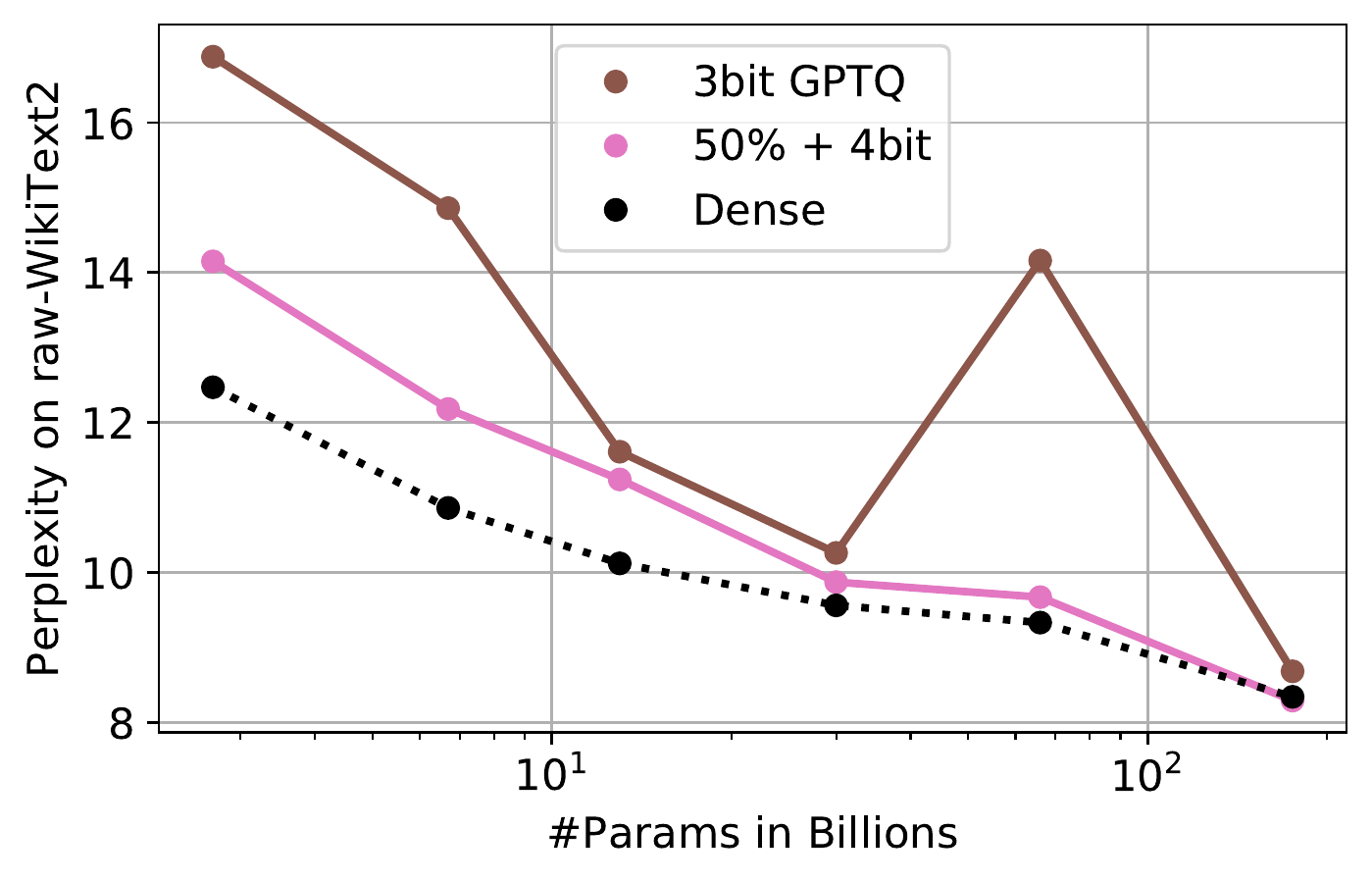}
    \vspace{-10pt}
    \caption{Comparing joint 50\% sparsity + 4-bit quantization with size-equivalent 3-bit on the OPT family for $\geq$ 2.7B params.}
    \label{fig:sparse-quantized}
\end{figure}

\paragraph{Joint Sparsification \& Quantization.} Another interesting research direction is the combination of sparsity and quantization, which would allow combining computational speedups from sparsity \cite{pmlr-v119-kurtz20a, elsen2020fast} with memory savings from quantization \cite{frantar2022gptq, dettmers2022llm, dettmers2022case}. Specifically, if we compress a model to 50\% sparse + 4-bit weights, store only the non-zero weights and use a bitmask to indicate their positions, then this has the same overall memory consumption as 3-bit quantization. Hence, in Figure \ref{fig:sparse-quantized} (right) we compare \sparsegpt{} 50\% + 4-bit with state-of-the-art GPTQ \cite{frantar2022gptq} 3-bit numbers. It can be seen that 50\% + 4-bit models are more accurate than their respective 3-bit versions for 2.7B+ parameter models, including 175B with 8.29 vs. 8.68 3-bit. We also tested 2:4 and 4:8 in combination with 4-bit on OPT-175B yielding 8.55 and 8.85 perplexities, suggesting that 4bit weight quantization only brings an $\approx 0.1$ perplexity increase on top semi-structured sparsity.

\begin{figure}[h!]
    \centering
    \includegraphics[width=\linewidth]{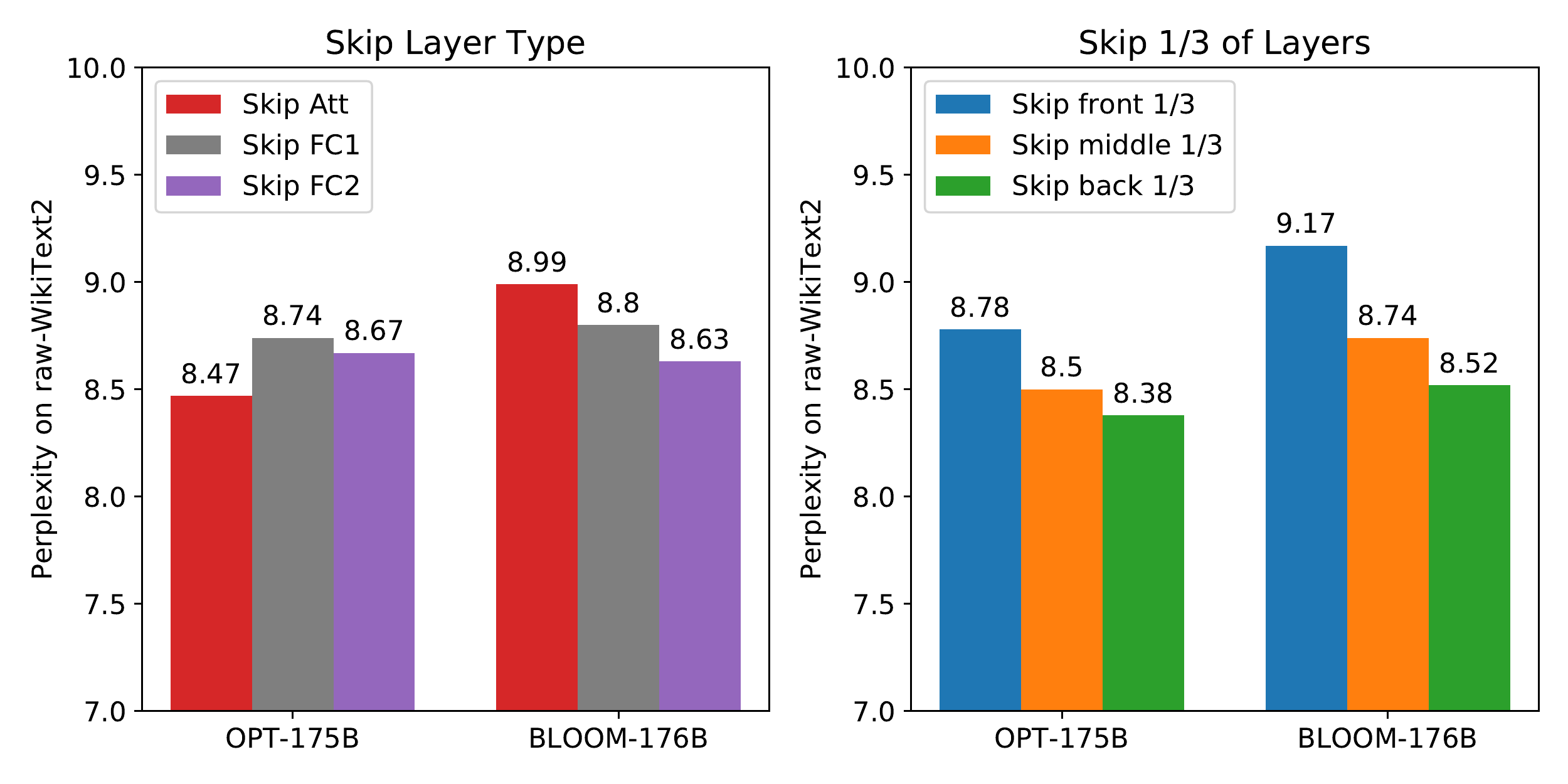}
    \vspace{-20pt}
    \caption{Sensitivity results for partial 2:4 pruning.}
    \label{fig:sensitivity}
\end{figure}

\paragraph{Sensitivity \& Partial N:M Sparsity.} One important practical question concerning n:m pruning is what to do when the fully sparsified model is not accurate enough? The overall sparsity level cannot simply be lowered uniformly, instead one must choose a subset of layers to n:m-sparsify completely. We now investigate what a good selection is in the context of extremely large language models: we assume that 2/3 of the layers of OPT-175B/BLOOM-176B should be pruned to 2:4 sparsity and consider skipping either all layers of one type (attention, fully-connected-1, fully-connected-2) or skipping one third of consecutive layers (front, middle, back). The results  are shown in Figure \ref{fig:sensitivity}.

While the sensitivity of layer-types differs noticeably between models, there appears to be a clear trend when it comes to model parts: \textit{later layers are more sensitive than earlier ones}; skipping the last third of the model gives the best accuracy. This has a very practical consequence in that, due to the sequential nature of \sparsegpt{}, we can generate a sequence of increasingly 2:4 sparsified models (e.g. 1/2, 2/3, 3/4, \dots) in \textit{a single pruning pass} by combining the first $x$ layers from a \sparsegpt{} run with the last $n_\text{layers} - x$ of the original model. The accuracy of such model sequences are shown in Appendix \ref{app:partial-24}.

\section{Related Work}

\textbf{Pruning Methods.} 
To our knowledge, we are the first to investigate pruning of massive GPT-scale models, e.g. with more than 10 billion parameters.
One justification for this surprising gap is the fact that most existing pruning methods, e.g.~\cite{han2015deep, gale2019state, kurtic2022gmp}, require \emph{extensive retraining} following the pruning step in order to recover accuracy, while GPT-scale models usually require massive amounts of computation and parameter tuning both for training or finetuning~\cite{zhang2022opt}.
\sparsegpt{} is a \emph{post-training} method for  GPT-scale models, as it does not perform any finetuning. 
So far, post-training pruning methods have only been investigated at the scale of classic CNN or BERT-type models~\cite{hubara2021accelerated, frantar2022obc, kwon2022fast}, which have 100-1000x fewer weights than our models of interest. 
We discussed the challenges of scaling these methods, and their relationship to \sparsegpt{}, in Section~\ref{sec:preliminaries}. 

\textbf{Post-Training Quantization.} 
By contrast, there has been significant work on post-training methods for \emph{quantizing} open GPT-scale models~\cite{zhang2022opt, scao2022bloom}. 
Specifically, the ZeroQuant~\cite{yao2022zeroquant}, LLM.int8()~\cite{dettmers2022llm} and nuQmm~\cite{park2022nuqmm} methods investigated the feasibility of round-to-nearest quantization for billion-parameter models, showing that 8-bit quantization for weights is feasible via this approach, but that activation quantization can be difficult due to the existence of outlier features. 
\citet{frantar2022gptq} leverage approximate second-order information for accurate quantization of weights down to 2--4 bits, for the very largest models, and show generative batch-size 1 inference speedups of 2-5x when coupled with efficient GPU kernels.
Follow-up work~\citep{xiao2022smoothquant} 
investigated joint activation and weight quantization  to 8 bits, proposing a smoothing-based scheme which reduces the difficulty of activation quantization and is complemented by efficient GPU kernels. 
\citet{park2022quadapter} tackle the hardness of quantizing activation outliers via \emph{quadapters}, learnable parameters whose goal is to scale activations channel-wise, while keeping the other model parameters unchanged. \citet{dettmers2022case} investigate scaling relationships between model size, quantization bits, and different notions of accuracy for massive LLMs, observing high correlations between perplexity scores and aggregated zero-shot accuracy across tasks.
As we have shown in Section~\ref{sec:joint}, the \sparsegpt{} algorithm can be applied in conjunction with GPTQ, the current state-of-the-art algorithm for weight quantization, and should be compatible with activation quantization approaches~\cite{xiao2022smoothquant, park2022quadapter}.

\section{Discussion}
\label{sec:discussion}

We have provided a new post-training pruning method called \sparsegpt{}, specifically tailored to massive language models from the GPT family. Our results show for the first time that large-scale generative pretrained Transformer-family models can be compressed to high sparsity via weight pruning in \emph{one-shot, without any retraining}, at low loss of accuracy, when measured both in terms of perplexity and zero-shot performance. 
Specifically, we have shown that the largest open-source GPT-family models (e.g. OPT-175B and BLOOM-176B)  can reach 50-60\% sparsity, dropping more than 100B weights, with low accuracy fluctuations.

Our work shows that the high degree of parametrization of massive GPT models allows pruning to directly identify sparse accurate models in the ``close neighborhood'' of the dense model, without gradient information. Remarkably, the output of such sparse models correlates extremely closely with that of the dense model. 
We also show that \emph{larger models are easier to sparsify}: at a fixed sparsity level, the relative accuracy drop for the larger sparse models narrows as we increase model size, to the point where inducing 50\% sparsity results in practically no accuracy decrease on the largest models, which should be seen as very encouraging for future work on compressing such massive models.  

\section{Acknowledgements}

The authors gratefully acknowledge funding from the European Research Council (ERC) under the European Union’s
Horizon 2020 programme (grant agreement No. 805223 ScaleML), as well as experimental support from Eldar Kurtic,
and from the IST Austria IT department, in particular Stefano Elefante, Andrei Hornoiu, and Alois Schloegl.

\bibliography{references}

\begin{thebibliography}{51}
\providecommand{\natexlab}[1]{#1}
\providecommand{\url}[1]{\texttt{#1}}
\expandafter\ifx\csname urlstyle\endcsname\relax
  \providecommand{\doi}[1]{doi: #1}\else
  \providecommand{\doi}{doi: \begingroup \urlstyle{rm}\Url}\fi

\bibitem[Blumensath \& Davies(2008)Blumensath and
  Davies]{blumensath2008iterative}
Blumensath, T. and Davies, M.~E.
\newblock Iterative thresholding for sparse approximations.
\newblock \emph{Journal of Fourier Analysis and Applications}, 14\penalty0
  (5-6):\penalty0 629--654, 2008.

\bibitem[Boratko et~al.(2018)Boratko, Padigela, Mikkilineni, Yuvraj, Das,
  McCallum, Chang, Fokoue-Nkoutche, Kapanipathi, Mattei,
  et~al.]{boratko2018systematic}
Boratko, M., Padigela, H., Mikkilineni, D., Yuvraj, P., Das, R., McCallum, A.,
  Chang, M., Fokoue-Nkoutche, A., Kapanipathi, P., Mattei, N., et~al.
\newblock A systematic classification of knowledge, reasoning, and context
  within the {ARC} dataset.
\newblock \emph{arXiv preprint arXiv:1806.00358}, 2018.

\bibitem[Dettmers \& Zettlemoyer(2022)Dettmers and
  Zettlemoyer]{dettmers2022case}
Dettmers, T. and Zettlemoyer, L.
\newblock The case for 4-bit precision: k-bit inference scaling laws.
\newblock \emph{arXiv preprint arXiv:2212.09720}, 2022.

\bibitem[Dettmers et~al.(2022)Dettmers, Lewis, Belkada, and
  Zettlemoyer]{dettmers2022llm}
Dettmers, T., Lewis, M., Belkada, Y., and Zettlemoyer, L.
\newblock {LLM.int8()}: 8-bit matrix multiplication for transformers at scale.
\newblock \emph{arXiv preprint arXiv:2208.07339}, 2022.

\bibitem[EleutherAI(2022)]{eleuther}
EleutherAI.
\newblock {EleutherAI LM Evaluation Harness}, 2022.
\newblock URL \url{https://github.com/EleutherAI/lm-evaluation-harness}.

\bibitem[Elsen et~al.(2020)Elsen, Dukhan, Gale, and Simonyan]{elsen2020fast}
Elsen, E., Dukhan, M., Gale, T., and Simonyan, K.
\newblock Fast sparse convnets.
\newblock In \emph{Conference on Computer Vision and Pattern Recognition
  (CVPR)}, 2020.

\bibitem[Evci et~al.(2020)Evci, Gale, Menick, Castro, and
  Elsen]{evci2020rigging}
Evci, U., Gale, T., Menick, J., Castro, P.~S., and Elsen, E.
\newblock Rigging the lottery: Making all tickets winners.
\newblock In \emph{International Conference on Machine Learning (ICML)}, 2020.

\bibitem[Frantar \& Alistarh(2022)Frantar and Alistarh]{frantar2022spdy}
Frantar, E. and Alistarh, D.
\newblock {SPDY:} {A}ccurate pruning with speedup guarantees.
\newblock \emph{arXiv preprint arXiv:2201.13096}, 2022.

\bibitem[Frantar et~al.(2021)Frantar, Kurtic, and Alistarh]{frantar2021m}
Frantar, E., Kurtic, E., and Alistarh, D.
\newblock {M-FAC}: Efficient matrix-free approximations of second-order
  information.
\newblock In \emph{Conference on Neural Information Processing Systems
  (NeurIPS)}, 2021.

\bibitem[Frantar et~al.(2022{\natexlab{a}})Frantar, Ashkboos, Hoefler, and
  Alistarh]{frantar2022gptq}
Frantar, E., Ashkboos, S., Hoefler, T., and Alistarh, D.
\newblock {GPTQ}: Accurate post-training compression for generative pretrained
  transformers.
\newblock \emph{arXiv preprint arXiv:2210.17323}, 2022{\natexlab{a}}.

\bibitem[Frantar et~al.(2022{\natexlab{b}})Frantar, Singh, and
  Alistarh]{frantar2022obc}
Frantar, E., Singh, S.~P., and Alistarh, D.
\newblock {Optimal Brain Compression}: A framework for accurate post-training
  quantization and pruning.
\newblock \emph{arXiv preprint arXiv:2208.11580}, 2022{\natexlab{b}}.
\newblock Accepted to NeurIPS 2022, to appear.

\bibitem[Gale et~al.(2019)Gale, Elsen, and Hooker]{gale2019state}
Gale, T., Elsen, E., and Hooker, S.
\newblock The state of sparsity in deep neural networks.
\newblock In \emph{International Conference on Machine Learning (ICML)}, 2019.

\bibitem[Hagiwara(1994)]{hagiwara1994}
Hagiwara, M.
\newblock A simple and effective method for removal of hidden units and
  weights.
\newblock \emph{Neurocomputing}, 6\penalty0 (2):\penalty0 207 -- 218, 1994.
\newblock ISSN 0925-2312.
\newblock Backpropagation, Part IV.

\bibitem[Han et~al.(2015)Han, Pool, Tran, and Dally]{han2015learning}
Han, S., Pool, J., Tran, J., and Dally, W.~J.
\newblock Learning both weights and connections for efficient neural networks.
\newblock In \emph{Conference on Neural Information Processing Systems
  (NeurIPS)}, 2015.

\bibitem[Han et~al.(2016)Han, Mao, and Dally]{han2015deep}
Han, S., Mao, H., and Dally, W.~J.
\newblock Deep compression: Compressing deep neural networks with pruning,
  trained quantization and {Huffman} coding.
\newblock In \emph{International Conference on Learning Representations
  (ICLR)}, 2016.

\bibitem[Hassibi et~al.(1993)Hassibi, Stork, and Wolff]{hassibi1993optimal}
Hassibi, B., Stork, D.~G., and Wolff, G.~J.
\newblock Optimal brain surgeon and general network pruning.
\newblock In \emph{IEEE International Conference on Neural Networks}, 1993.

\bibitem[He et~al.(2018)He, Lin, Liu, Wang, Li, and Han]{he2018amc}
He, Y., Lin, J., Liu, Z., Wang, H., Li, L.-J., and Han, S.
\newblock {AMC}: {AutoML} for model compression and acceleration on mobile
  devices.
\newblock In \emph{European Conference on Computer Vision (ECCV)}, 2018.

\bibitem[Hoefler et~al.(2021)Hoefler, Alistarh, Ben-Nun, Dryden, and
  Peste]{hoefler2021sparsity}
Hoefler, T., Alistarh, D., Ben-Nun, T., Dryden, N., and Peste, A.
\newblock Sparsity in deep learning: Pruning and growth for efficient inference
  and training in neural networks.
\newblock \emph{arXiv preprint arXiv:2102.00554}, 2021.

\bibitem[Hubara et~al.(2021{\natexlab{a}})Hubara, Chmiel, Island, Banner, Naor,
  and Soudry]{hubara2021accelerated}
Hubara, I., Chmiel, B., Island, M., Banner, R., Naor, S., and Soudry, D.
\newblock Accelerated sparse neural training: A provable and efficient method
  to find {N:M} transposable masks.
\newblock In \emph{Conference on Neural Information Processing Systems
  (NeurIPS)}, 2021{\natexlab{a}}.

\bibitem[Hubara et~al.(2021{\natexlab{b}})Hubara, Nahshan, Hanani, Banner, and
  Soudry]{hubara2021accurate}
Hubara, I., Nahshan, Y., Hanani, Y., Banner, R., and Soudry, D.
\newblock Accurate post training quantization with small calibration sets.
\newblock In \emph{International Conference on Machine Learning (ICML)},
  2021{\natexlab{b}}.

\bibitem[HuggingFace(2022)]{hfperplexity}
HuggingFace.
\newblock {HuggingFace Perplexity Calculation}, 2022.
\newblock URL \url{https://huggingface.co/docs/transformers/perplexity}.

\bibitem[Kingdon(1997)]{Kingdon1997}
Kingdon, J.
\newblock \emph{Hypothesising Neural Nets}, pp.\  81--106.
\newblock Springer London, London, 1997.
\newblock ISBN 978-1-4471-0949-5.
\newblock \doi{10.1007/978-1-4471-0949-5_5}.

\bibitem[Kurtic \& Alistarh(2022)Kurtic and Alistarh]{kurtic2022gmp}
Kurtic, E. and Alistarh, D.
\newblock Gmp*: Well-tuned global magnitude pruning can outperform most
  bert-pruning methods.
\newblock \emph{arXiv preprint arXiv:2210.06384}, 2022.

\bibitem[Kurtic et~al.(2022)Kurtic, Campos, Nguyen, Frantar, Kurtz, Fineran,
  Goin, and Alistarh]{kurtic2022optimal}
Kurtic, E., Campos, D., Nguyen, T., Frantar, E., Kurtz, M., Fineran, B., Goin,
  M., and Alistarh, D.
\newblock The {Optimal BERT Surgeon}: Scalable and accurate second-order
  pruning for large language models.
\newblock \emph{arXiv preprint arXiv:2203.07259}, 2022.

\bibitem[Kurtz et~al.(2020)Kurtz, Kopinsky, Gelashvili, Matveev, Carr, Goin,
  Leiserson, Moore, Nell, Shavit, and Alistarh]{pmlr-v119-kurtz20a}
Kurtz, M., Kopinsky, J., Gelashvili, R., Matveev, A., Carr, J., Goin, M.,
  Leiserson, W., Moore, S., Nell, B., Shavit, N., and Alistarh, D.
\newblock Inducing and exploiting activation sparsity for fast inference on
  deep neural networks.
\newblock In \emph{International Conference on Machine Learning (ICML)}, 2020.

\bibitem[Kwon et~al.(2022)Kwon, Kim, Mahoney, Hassoun, Keutzer, and
  Gholami]{kwon2022fast}
Kwon, W., Kim, S., Mahoney, M.~W., Hassoun, J., Keutzer, K., and Gholami, A.
\newblock A fast post-training pruning framework for transformers.
\newblock \emph{arXiv preprint arXiv:2204.09656}, 2022.

\bibitem[LeCun et~al.(1989)LeCun, Denker, and Solla]{lecun1990optimal}
LeCun, Y., Denker, J.~S., and Solla, S.~A.
\newblock Optimal brain damage.
\newblock In \emph{Conference on Neural Information Processing Systems
  (NeurIPS)}, 1989.

\bibitem[Li et~al.(2021)Li, Gong, Tan, Yang, Hu, Zhang, Yu, Wang, and
  Gu]{li2021brecq}
Li, Y., Gong, R., Tan, X., Yang, Y., Hu, P., Zhang, Q., Yu, F., Wang, W., and
  Gu, S.
\newblock {BRECQ}: Pushing the limit of post-training quantization by block
  reconstruction.
\newblock In \emph{International Conference on Learning Representations
  (ICLR)}, 2021.

\bibitem[Liu et~al.(2021)Liu, Zhang, Kuang, Zhou, Xue, Wang, Chen, Yang, Liao,
  and Zhang]{liu2021group}
Liu, L., Zhang, S., Kuang, Z., Zhou, A., Xue, J.-H., Wang, X., Chen, Y., Yang,
  W., Liao, Q., and Zhang, W.
\newblock Group fisher pruning for practical network compression.
\newblock In \emph{International Conference on Machine Learning (ICML)}, 2021.

\bibitem[Marcus et~al.(1994)Marcus, Kim, Marcinkiewicz, MacIntyre, Bies,
  Ferguson, Katz, and Schasberger]{PTB}
Marcus, M., Kim, G., Marcinkiewicz, M.~A., MacIntyre, R., Bies, A., Ferguson,
  M., Katz, K., and Schasberger, B.
\newblock The penn treebank: Annotating predicate argument structure.
\newblock In \emph{Human Language Technology: Proceedings of a Workshop held at
  Plainsboro, New Jersey, March 8-11, 1994}, 1994.

\bibitem[Merity et~al.(2016)Merity, Xiong, Bradbury, and Socher]{wikitext103}
Merity, S., Xiong, C., Bradbury, J., and Socher, R.
\newblock Pointer sentinel mixture models.
\newblock \emph{arXiv preprint arXiv:1609.07843}, 2016.

\bibitem[Mishra et~al.(2021)Mishra, Latorre, Pool, Stosic, Stosic, Venkatesh,
  Yu, and Micikevicius]{NVIDIASparse}
Mishra, A., Latorre, J.~A., Pool, J., Stosic, D., Stosic, D., Venkatesh, G.,
  Yu, C., and Micikevicius, P.
\newblock Accelerating sparse deep neural networks.
\newblock \emph{arXiv preprint arXiv:2104.08378}, 2021.

\bibitem[Mostafazadeh et~al.(2017)Mostafazadeh, Roth, Louis, Chambers, and
  Allen]{mostafazadeh2017lsdsem}
Mostafazadeh, N., Roth, M., Louis, A., Chambers, N., and Allen, J.
\newblock Lsdsem 2017 shared task: The story cloze test.
\newblock In \emph{Proceedings of the 2nd Workshop on Linking Models of
  Lexical, Sentential and Discourse-level Semantics}, pp.\  46--51, 2017.

\bibitem[Nagel et~al.(2020)Nagel, Amjad, Van~Baalen, Louizos, and
  Blankevoort]{nagel2020up}
Nagel, M., Amjad, R.~A., Van~Baalen, M., Louizos, C., and Blankevoort, T.
\newblock Up or down? {A}daptive rounding for post-training quantization.
\newblock In \emph{International Conference on Machine Learning (ICML)}, 2020.

\bibitem[NeuralMagic(2022)]{deepsparse}
NeuralMagic.
\newblock {DeepSparse}, 2022.
\newblock URL \url{https://github.com/neuralmagic/deepsparse}.

\bibitem[Paperno et~al.(2016)Paperno, Kruszewski, Lazaridou, Pham, Bernardi,
  Pezzelle, Baroni, Boleda, and Fern{\'a}ndez]{paperno2016lambada}
Paperno, D., Kruszewski, G., Lazaridou, A., Pham, Q.~N., Bernardi, R.,
  Pezzelle, S., Baroni, M., Boleda, G., and Fern{\'a}ndez, R.
\newblock The {LAMBADA} dataset: Word prediction requiring a broad discourse
  context.
\newblock \emph{arXiv preprint arXiv:1606.06031}, 2016.

\bibitem[Park et~al.(2022{\natexlab{a}})Park, Park, Kwon, Kim, Lee, and
  Lee]{park2022nuqmm}
Park, G., Park, B., Kwon, S.~J., Kim, B., Lee, Y., and Lee, D.
\newblock {nuQmm}: Quantized matmul for efficient inference of large-scale
  generative language models.
\newblock \emph{arXiv preprint arXiv:2206.09557}, 2022{\natexlab{a}}.

\bibitem[Park et~al.(2022{\natexlab{b}})Park, You, Nagel, and
  Chang]{park2022quadapter}
Park, M., You, J., Nagel, M., and Chang, S.
\newblock Quadapter: Adapter for gpt-2 quantization.
\newblock \emph{arXiv preprint arXiv:2211.16912}, 2022{\natexlab{b}}.

\bibitem[Paszke et~al.(2019)Paszke, Gross, Massa, Lerer, Bradbury, Chanan,
  Killeen, Lin, Gimelshein, Antiga, et~al.]{paszke2019pytorch}
Paszke, A., Gross, S., Massa, F., Lerer, A., Bradbury, J., Chanan, G., Killeen,
  T., Lin, Z., Gimelshein, N., Antiga, L., et~al.
\newblock Pytorch: An imperative style, high-performance deep learning library.
\newblock In \emph{Conference on Neural Information Processing Systems
  (NeurIPS)}, 2019.

\bibitem[Peste et~al.(2021)Peste, Iofinova, Vladu, and Alistarh]{peste2021ac}
Peste, A., Iofinova, E., Vladu, A., and Alistarh, D.
\newblock {AC/DC}: Alternating compressed/decompressed training of deep neural
  networks.
\newblock In \emph{Conference on Neural Information Processing Systems
  (NeurIPS)}, 2021.

\bibitem[Raffel et~al.(2020)Raffel, Shazeer, Roberts, Lee, Narang, Matena,
  Zhou, Li, and Liu]{C4}
Raffel, C., Shazeer, N., Roberts, A., Lee, K., Narang, S., Matena, M., Zhou,
  Y., Li, W., and Liu, P.
\newblock Exploring the limits of transfer learning with a unified text-to-text
  transformer.
\newblock \emph{Journal of Machine Learning Research}, 21\penalty0
  (140):\penalty0 1--67, 2020.

\bibitem[Sanh et~al.(2020)Sanh, Wolf, and Rush]{2020-sanh}
Sanh, V., Wolf, T., and Rush, A.~M.
\newblock Movement pruning: Adaptive sparsity by fine-tuning.
\newblock \emph{arXiv preprint arXiv:2005.07683}, 2020.

\bibitem[Scao et~al.(2022)Scao, Fan, Akiki, Pavlick, Ili{\'c}, Hesslow,
  Castagn{\'e}, Luccioni, Yvon, Gall{\'e}, et~al.]{scao2022bloom}
Scao, T.~L., Fan, A., Akiki, C., Pavlick, E., Ili{\'c}, S., Hesslow, D.,
  Castagn{\'e}, R., Luccioni, A.~S., Yvon, F., Gall{\'e}, M., et~al.
\newblock Bloom: A 176b-parameter open-access multilingual language model.
\newblock \emph{arXiv preprint arXiv:2211.05100}, 2022.

\bibitem[Singh \& Alistarh(2020)Singh and Alistarh]{singh2020woodfisher}
Singh, S.~P. and Alistarh, D.
\newblock {WoodFisher}: Efficient second-order approximation for neural network
  compression.
\newblock In \emph{Conference on Neural Information Processing Systems
  (NeurIPS)}, 2020.

\bibitem[Tata \& Patel(2003)Tata and Patel]{tata2003piqa}
Tata, S. and Patel, J.~M.
\newblock {PiQA}: An algebra for querying protein data sets.
\newblock In \emph{International Conference on Scientific and Statistical
  Database Management}, 2003.

\bibitem[Wolf et~al.(2019)Wolf, Debut, Sanh, Chaumond, Delangue, Moi, Cistac,
  Rault, Louf, Funtowicz, et~al.]{wolf2019huggingface}
Wolf, T., Debut, L., Sanh, V., Chaumond, J., Delangue, C., Moi, A., Cistac, P.,
  Rault, T., Louf, R., Funtowicz, M., et~al.
\newblock Huggingface's transformers: State-of-the-art natural language
  processing.
\newblock \emph{arXiv preprint arXiv:1910.03771}, 2019.

\bibitem[Xiao et~al.(2022)Xiao, Lin, Seznec, Demouth, and
  Han]{xiao2022smoothquant}
Xiao, G., Lin, J., Seznec, M., Demouth, J., and Han, S.
\newblock Smoothquant: Accurate and efficient post-training quantization for
  large language models.
\newblock \emph{arXiv preprint arXiv:2211.10438}, 2022.

\bibitem[Yao et~al.(2022)Yao, Aminabadi, Zhang, Wu, Li, and
  He]{yao2022zeroquant}
Yao, Z., Aminabadi, R.~Y., Zhang, M., Wu, X., Li, C., and He, Y.
\newblock {ZeroQuant}: Efficient and affordable post-training quantization for
  large-scale transformers.
\newblock \emph{arXiv preprint arXiv:2206.01861}, 2022.

\bibitem[Zhang et~al.(2022)Zhang, Roller, Goyal, Artetxe, Chen, Chen, Dewan,
  Diab, Li, Lin, et~al.]{zhang2022opt}
Zhang, S., Roller, S., Goyal, N., Artetxe, M., Chen, M., Chen, S., Dewan, C.,
  Diab, M., Li, X., Lin, X.~V., et~al.
\newblock {OPT}: Open pre-trained transformer language models.
\newblock \emph{arXiv preprint arXiv:2205.01068}, 2022.

\bibitem[Zhou et~al.(2021)Zhou, Ma, Zhu, Liu, Zhang, Yuan, Sun, and
  Li]{zhou2021learning}
Zhou, A., Ma, Y., Zhu, J., Liu, J., Zhang, Z., Yuan, K., Sun, W., and Li, H.
\newblock Learning {N:M} fine-grained structured sparse neural networks from
  scratch.
\newblock In \emph{International Conference on Learning Representations
  (ICLR)}, 2021.

\bibitem[Zhu \& Gupta(2017)Zhu and Gupta]{zhu2017prune}
Zhu, M. and Gupta, S.
\newblock To prune, or not to prune: exploring the efficacy of pruning for
  model compression.
\newblock \emph{arXiv preprint arXiv:1710.01878}, 2017.

\end{thebibliography}
\bibliographystyle{arxiv}


\newpage
\appendix
\onecolumn

\section{Ablation Studies}
\label{app:ablations}

In this section, we conduct ablations studies with respect to several of the main parameters of \sparsegpt{}. For a fast iteration time and making it possible to also explore more compute and memory intensive settings, we focus on the OPT-2.7B model here. Unless stated otherwise, we always prune uniformly to the default 50\% sparsity. For brevity we only show raw-WikiText2 results here, but would like to note that the behavior on other datasets is very similar.

\begin{figure*}[h]
    \centering
    \begin{minipage}[c]{.32\textwidth}
        \centering
        \includegraphics[width=\linewidth]{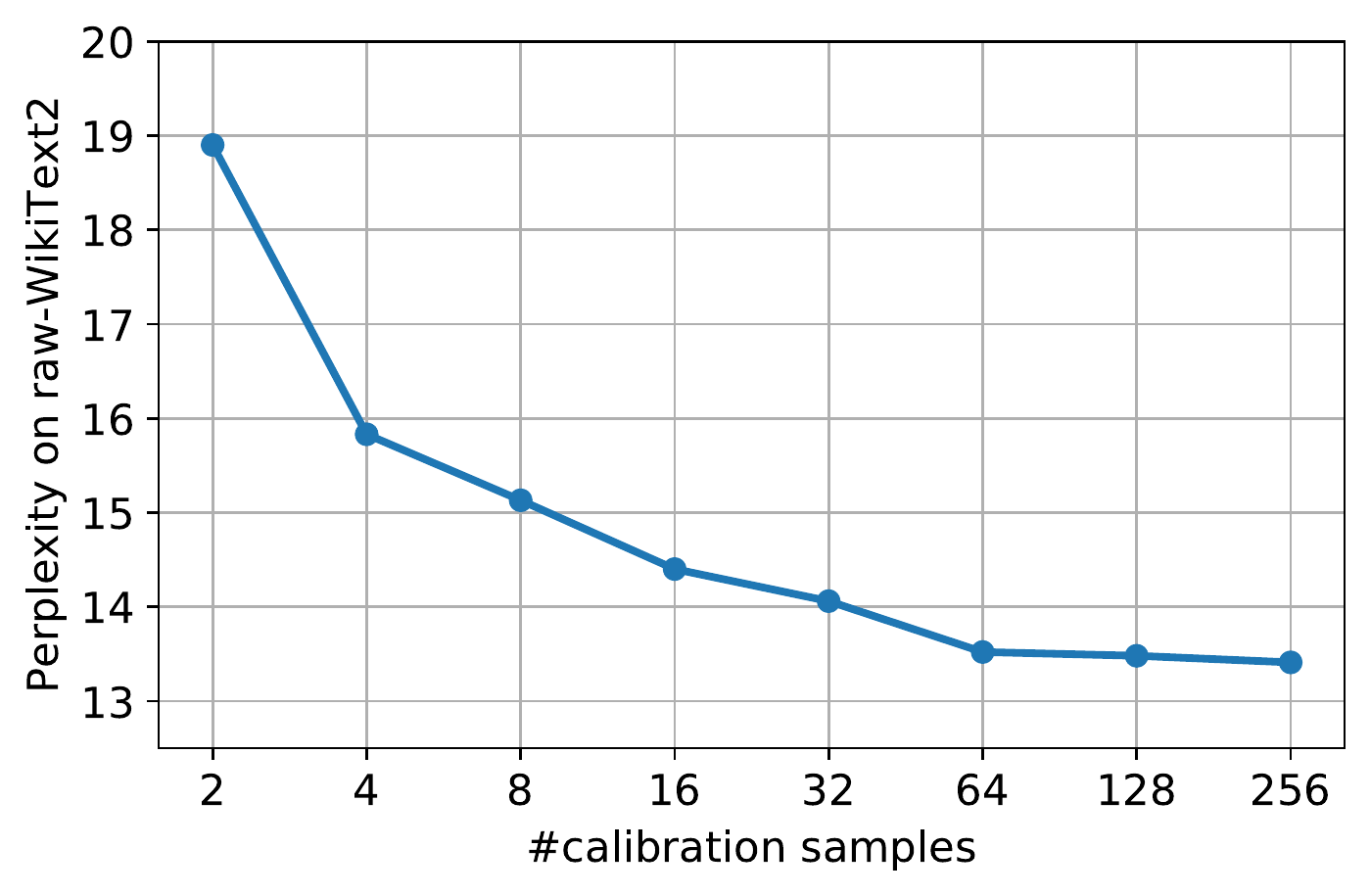}
        \vspace{-20pt}
        \captionof{figure}{Calibration samples ablation.}
        \label{fig:ablation-samples}
    \end{minipage}
    \begin{minipage}[c]{.32\textwidth}
        \centering
        \includegraphics[width=\linewidth]{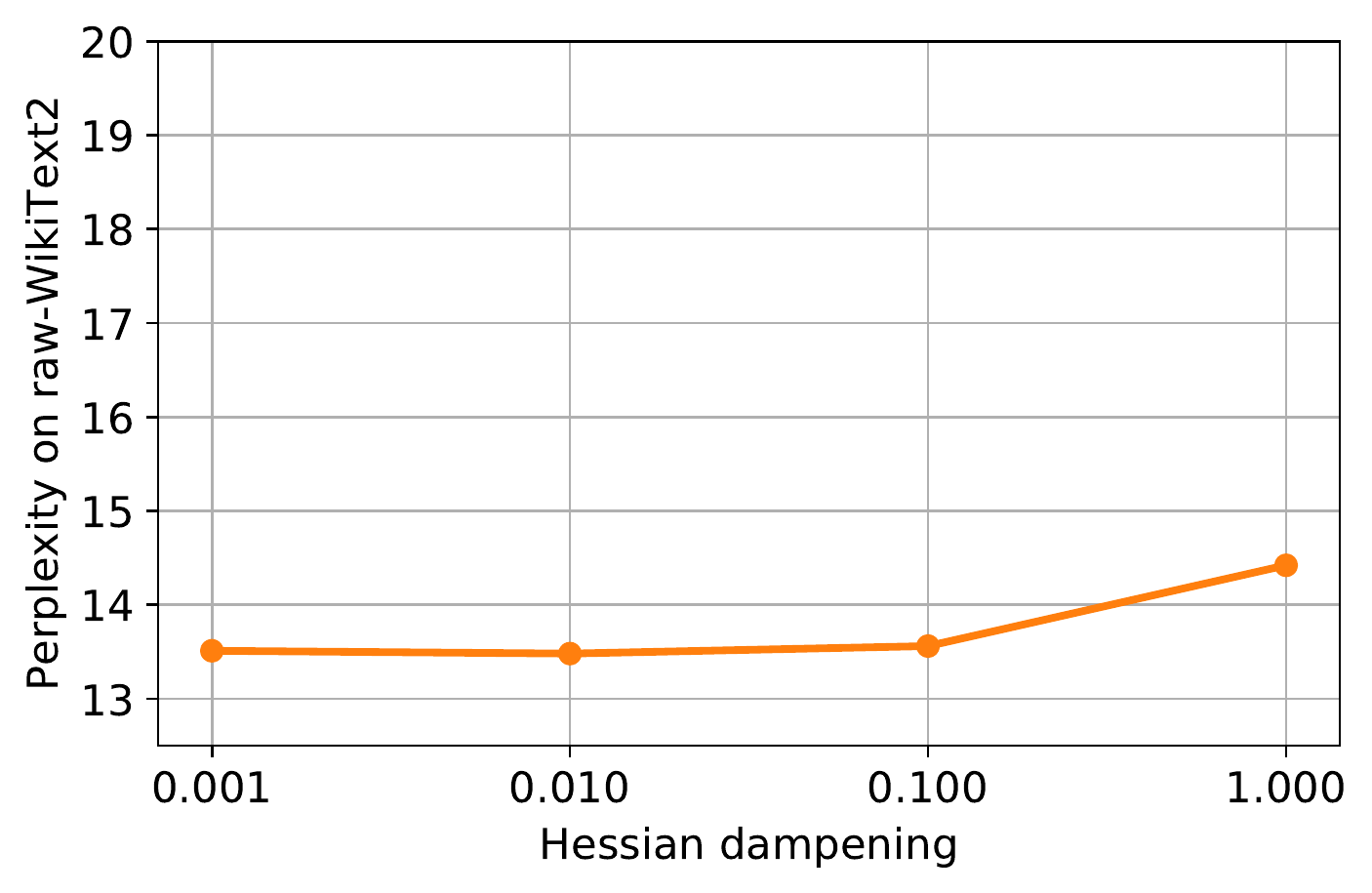}
        \vspace{-20pt}
        \captionof{figure}{Hessian dampening ablation.}
        \label{fig:ablation-dampening}
    \end{minipage}
    \begin{minipage}[c]{.32\textwidth}
        \centering
        \includegraphics[width=\linewidth]{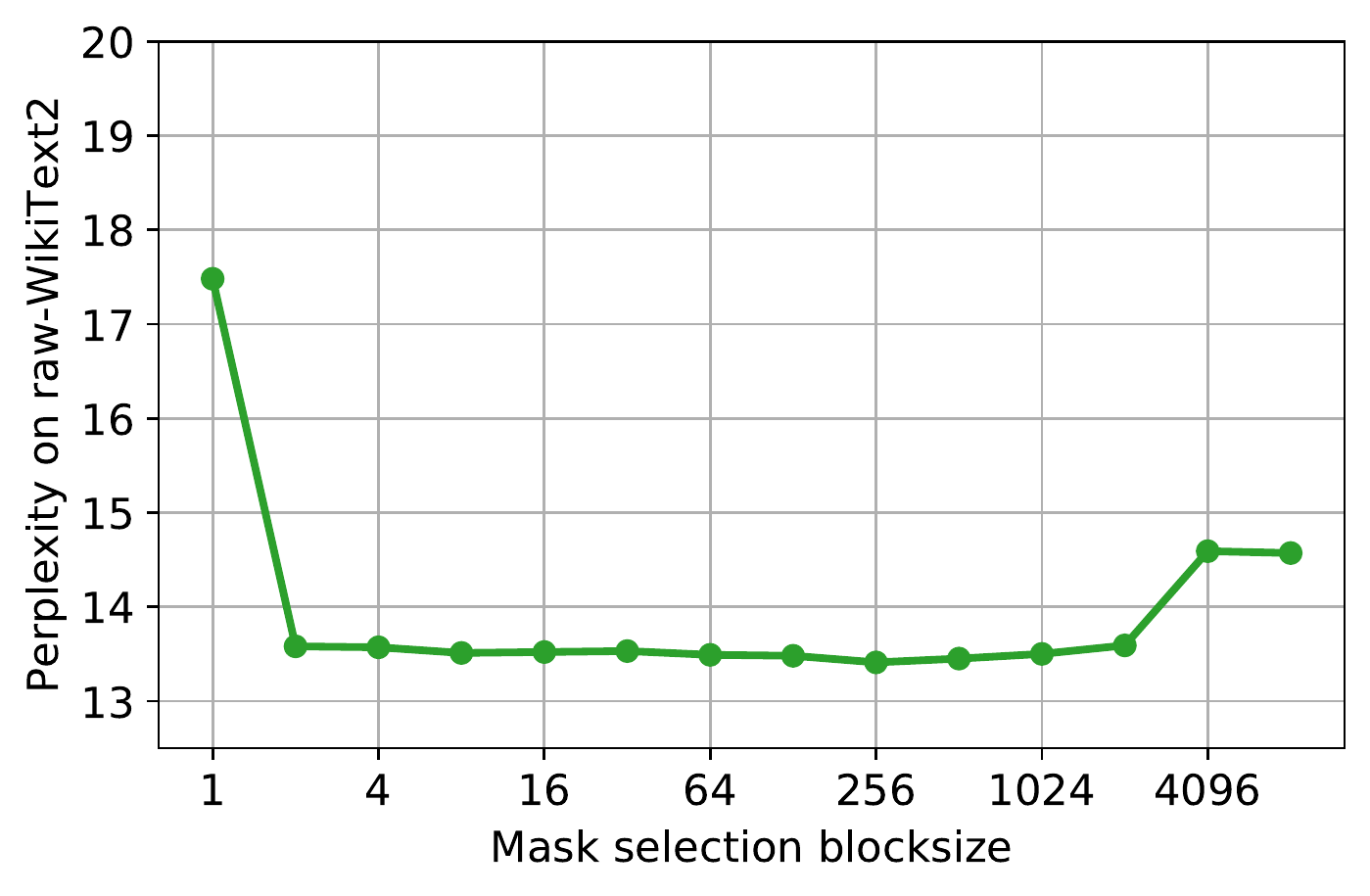}
        \vspace{-20pt}
        \captionof{figure}{Mask select. blocksize ablation.}
        \label{fig:ablation-blocksize}
    \end{minipage}
\end{figure*}

\paragraph{Amount of Calibration Data.} First, we investigate how the accuracy of \sparsegpt{} scales with the number calibration data samples, which we vary in powers of two. The results are shown in Figure \ref{fig:ablation-samples}. Curiously, \sparsegpt{} is already able to achieve decent results even with just a few 2048-token segments; using more samples however yields significant further improvements, but only up to a certain point as the curve flattens quite quickly. Thus, since using more samples also increases compute and memory costs, we stick to 128 samples in all our experiments.

\paragraph{Hessian Dampening.} Next, we study the impact of Hessian dampening by testing values varying as powers of ten (see Figure \ref{fig:ablation-dampening}) which are multiplied by the average diagonal value, following \cite{frantar2022gptq}. Overall, this parameter does not seem to be too sensitive, $0.001$ to $0.1$ appear to perform quite similar; only when the dampening is very high, the solution quality decreases significantly. We choose $1\%$ (i.e. $0.01$) dampening to be on the safe side with respect to inverse calculations also for the very largest models.

\paragraph{Mask Selection Blocksize.} Another important component of our method is the adaptive mask selection as shown in Figure~\ref{fig:ablation-blocksize} where we vary the corresponding blocksize parameter with powers of two. Both column-wise (blocksize 1) as well as near full blocksize (4096 and 8192) perform significantly worse than reasonable blocking. Interestingly, a wide range of block-sizes appear to work well, with ones around a few hundred being very slightly more accurate. We thus choose blocksize 128 which lies in that range while also slightly simplifying the algorithm implementation as it matches the default lazy weight update batchsize.

\paragraph{Sensitivity to Random Seeds.} Finally, we determine how sensitive the results of our algorithm are with respect to randomness; specifically, relative to the random sampling of the calibration data. We repeat a standard 50\% pruning run 5 times with different random seeds for data sampling and get $13.52 \pm 0.075$ (mean/std) suggesting that \sparsegpt{} is quite robust to the precise calibration data being used, which is in line with the observations in other post-training works \cite{nagel2020up, hubara2021accurate, frantar2022obc}.

\subsection{Approximation Quality}
\label{sec:approx-quality}

\begin{figure}[h!]
    \centering
    \includegraphics[width=.9\textwidth]{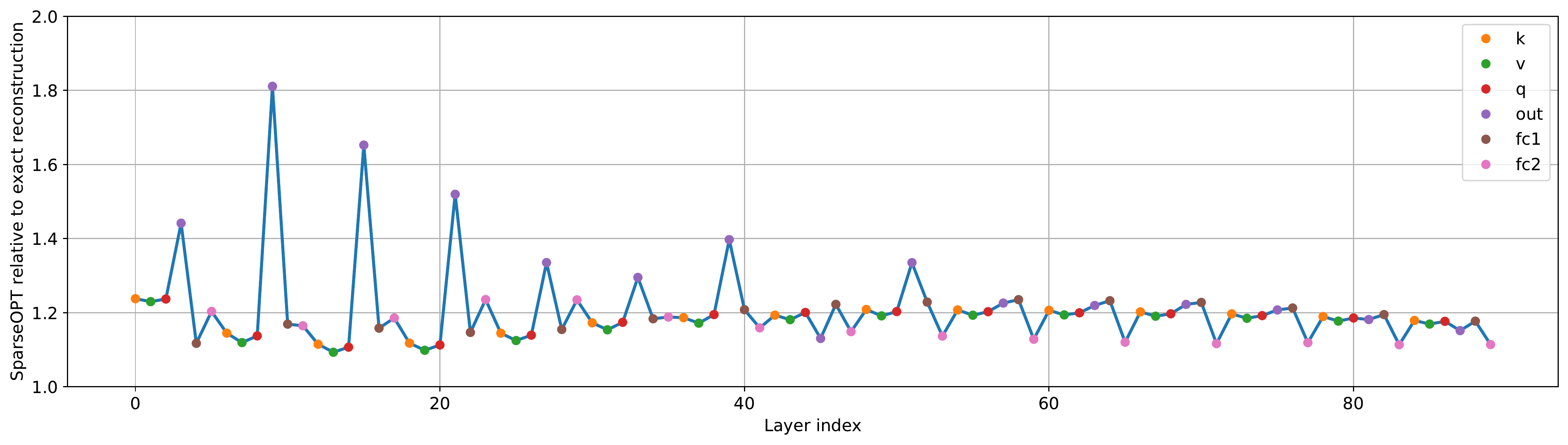}
    \vspace{-10pt}
    \caption{Error of \sparsegpt{} reconstruction relative to exact reconstruction for the first half of OPT-2.7B at 50\% sparsity.}
    \label{fig:errors}
\end{figure}

In this section we investigate how much is lost by the partial-update approximation employed by \sparsegpt{}, relative to (much more expensive) exact reconstruction. We again consider the OPT-2.7B model at 50\% sparsity and plot the layer-wise squared error of \sparsegpt{} relative to the error of exact reconstruction (with the same mask and Hessian) for the first half of the model in Figure \ref{fig:errors}. Apart from some outliers in form of the early attention out-projection layers, the final reconstruction errors of \sparsegpt{} seem to be on average only around 20\% worse than exact reconstruction; on the later fully-connected-2 layers, the approximation error even gets close to only 10\%, presumably because these layers have a very large number of total inputs and thus losses by considering only correlations within subsets are less severe than on smaller layers. Overall, these results suggest that, despite its dramatic speedup, \sparsegpt{} also remains quite accurate.

\section{Evaluation Details}
\label{app:evaluation-details}

\paragraph{Perplexity.} As mentioned in the main text, our perplexity calculation is carried out in standard fashion, following exactly the description of \cite{hfperplexity}. Concretely, that means we concatenate all samples in the test/validation dataset, encode the result with the model's matching tokenizer and then split it into non-overlapping segments of 2048 tokens (the maximum history of the models we study). Those are run through the model to calculate the corresponding average language modelling loss. The exponentiated number is the perplexity we report.

\paragraph{Datasets.} In terms of datasets, we use the raw version of the WikiText2 test-set and concatenate samples, as recommended by the HuggingFace description referenced above, with ``\textbackslash n\textbackslash n" to produce properly formatted markdown. For PTB, we use the test-set of HuggingFace's ``ptb\_text\_only'' version and concatenate samples directly, without separators, as PTB is not supposed to contain any punctuation. Our C4 subset consists of the starting (the dataset comes in random order) 256 times 2048 encoded tokens in the first shard of the directly concatenated validation set; this choice is made to keep evaluation costs manageable.

\section{Additional Results}
\label{app:additional-experiments}

\paragraph{Pruning Difficulty Scaling on PTB \& C4.} Tables \ref{tab:opt-ptb} and \ref{tab:opt-c4} present the equivalent results to Table \ref{tab:opt-wikitext2} in the main text, but on PTB and our C4 subset, respectively. Overall, they follow very similar trends to those discussed in Section \ref{sec:results}. The main notable difference is that no slight perplexity decrease relative to the dense baseline is observed at 50\% sparsity for the largest models, hence we have labelled this as a dataset specific phenomenon.
\vspace{-5pt}
\begin{table*}[h!]
    \centering
    \caption{OPT perplexity results on PTB.}
    \vspace{5pt}
    \scalebox{.9}{
        \begin{tabular}{|l|c|c|c|}
            \toprule
            OPT - 50\% & 125M & 350M & 1.3B \\
            \midrule
            Dense & 38.99 & 31.07 & 20.29 \\
            \midrule
            Magnitude & 276. & 126. & 3.1e3 \\
            AdaPrune & 92.14 & 64.64 & 41.60 \\
            \sparsegpt{} & \textbf{55.06} &\textbf{43.80} & \textbf{25.80} \\
            \bottomrule
        \end{tabular}
    }
    \scalebox{.9}{
        \begin{tabular}{|l|c|c|c|c|c|c|c|}
            \toprule
            OPT & Sparsity & 2.7B & 6.7B & 13B & 30B & 66B & 175B \\
            \midrule
            Dense & 0\% & 17.97 & 15.77 & 14.52 & 14.04 & 13.36 & 12.01 \\
            \midrule
            Magnitude & 50\% & 262. & 613. & 1.8e4 & 221. & 4.0e3 & 2.3e3 \\
            \sparsegpt{} & 50\% & \textbf{20.45} & \textbf{17.44} & \textbf{15.97} & \textbf{14.98} & \textbf{14.15} & \textbf{12.37} \\
            \midrule
            \sparsegpt{} & 4:8 & 23.02 & 18.84 & 17.23 & 15.68 & 14.68 & 12.78 \\
            \sparsegpt{} & 2:4 & 26.88 & 21.57 & 18.71 & 16.62 & 15.41 & 13.24 \\
            \bottomrule
        \end{tabular}
    }
    \label{tab:opt-ptb}
\end{table*}
\vspace{-15pt}
\begin{table*}[h!]
    \centering
    \caption{OPT perplexity results on a C4 subset.}
    \vspace{5pt}
    \scalebox{.9}{
        \begin{tabular}{|l|c|c|c|}
            \toprule
            OPT - 50\% & 125M & 350M & 1.3B \\
            \midrule
            Dense & 26.56 & 22.59 & 16.07 \\
            \midrule
            Magnitude & 141. & 77.04 & 403. \\
            AdaPrune & 48.84 & 39.15 & 28.56 \\
            \sparsegpt{} & \textbf{33.42} &\textbf{29.18} & \textbf{19.36} \\
            \bottomrule
        \end{tabular}
    }
    \scalebox{.9}{
        \begin{tabular}{|l|c|c|c|c|c|c|c|}
            \toprule
            OPT & Sparsity & 2.7B & 6.7B & 13B & 30B & 66B & 175B \\
            \midrule
            Dense & 0\% & 14.32 & 12.71 & 12.06 & 11.45 & 10.99 & 10.13 \\
            \midrule
            Magnitude & 50\% & 63.43 & 334. & 1.1e4 & 98.49 & 2.9e3 & 1.7e3 \\
            \sparsegpt{} & 50\% & \textbf{15.78} & \textbf{13.73} & \textbf{12.97} & \textbf{11.97} & \textbf{11.41} & \textbf{10.36} \\
            \midrule
            \sparsegpt{} & 4:8 & 17.21 & 14.77 & 13.76 & 12.48 & 11.77 & 10.61 \\
            \sparsegpt{} & 2:4 & 19.36 & 16.40 & 14.85 & 13.17 & 12.25 & 10.92 \\
            \bottomrule
        \end{tabular}
    }
    \label{tab:opt-c4}
\end{table*}

\paragraph{50\% Sparse + 3-bit.} The main paper only presents near loss-less results for 50\% + 4-bit joint sparsification and quantization, corresponding to 3-bit quantization in terms of storage. For 50\% + 3-bit (corresponding to 2.5-bit), OPT-175B achieves 8.60 PPL on raw-WikiText2, which is also more accurate than GPTQ's \cite{frantar2022gptq} 8.94 state-of-the-art 2.5-bit result. \sparsegpt{} scores the same 8.93 for 4:8 + 3-bit. Based on these initial investigations, we believe that combining sparsity + quantization is a promising direction towards even more extreme compression of very large language models.

\section{Partial 2:4 Results}
\label{app:partial-24}

Tables \ref{tab:opt-part24} and \ref{tab:bloom-part24} show the performance of a sequence of partially 2:4 sparse models on three different language modelling datasets. The first fraction of layers is fully sparsified while the remainder is kept dense. In this way, speedup and accuracy can be traded off also from binary compression choices, such as n:m-pruning.
\vspace{-5pt}
\begin{table}[h!]
    \centering
    \caption{Pruning different fractions (as consecutive segments from the beginning) of OPT-175B layers to the 2:4 pattern.}
    \vspace{5pt}
    \scalebox{.9}{
        \begin{tabular}{|l|c|c|c|c|c|c|}
            \toprule
            OPT-175B -- 2:4 & dense & 1/2 & 2/3 & 3/4 & 4/5 & full \\
            \midrule
            raw-WikiText2 & 8.34 & 8.22 & 8.38 & 8.49 & 8.52 & 8.74 \\
            PTB & 12.01 & 12.15 & 12.80 & 13.02 & 13.12 & 13.25 \\
            C4-subset & 10.13 & 10.22 & 10.41 & 10.52 & 10.59 & 10.92 \\
            \bottomrule
        \end{tabular}
    }
    \label{tab:opt-part24}
\end{table}
\vspace{-15pt}
\begin{table}[h!]
    \centering
    \caption{Pruning different fractions (as consecutive segments from the beginning) of BLOOM-176B layers to the 2:4 pattern.}
    \vspace{5pt}
    \scalebox{.9}{
        \begin{tabular}{|l|c|c|c|c|c|c|}
            \toprule
            BLOOM-176B -- 2:4 & dense & 1/2 & 2/3 & 3/4 & 4/5 & full \\
            \midrule
            raw-WikiText2 & 8.11 & 8.20 & 8.50 & 8.67 & 8.74 & 9.20 \\
            PTB & 14.58 & 14.78 & 15.44 & 15.84 & 15.96 & 16.42 \\
            C4-subset & 11.71 & 11.81 & 12.06 & 12.23 & 12.32 & 12.67 \\
            \bottomrule
        \end{tabular}
    }
    \label{tab:bloom-part24}
\end{table}

\section{Sparsity Acceleration}
\label{app:speedup}

Lastly, we perform a preliminary study of how well sparse language models can already be accelerated in practice with off-the-shelf tools, for both CPU and GPU inference. We think that these results can likely be improved significantly with more model specific optimization, which we think is an important topic for future work.

\paragraph{CPU Speedups.} First, we investigate acceleration of \textit{unstructured} sparsity for CPU inference. For that we utilize the state-of-the-art DeepSparse engine \cite{deepsparse} and run end-to-end inference on OPT-2.7B (support for larger variants appears to be still under development) for a single batch of 400 tokens, on an Intel(R) Core(TM) i9-7980XE CPU @ 2.60GHz using 18 cores. Table~\ref{tab:cpu-inference} shows the end-to-end speedups of running sparse models over the dense one, executed in the same engine/environment. (For reference, dense DeepSparse is $1.5\times$ faster than the standard ONNXRuntime.) The achieved speedups are close to the theoretical optimum, which suggests that unstructured sparsity acceleration for LLM inference on CPUs is already quite practical.

\begin{table}[h!]
    \centering
    \begin{tabular}{|l|c|c|c|}
        \toprule
        Sparsity & 40\% & 50\% & 60\%  \\
        \midrule
        Speedup & $1.57\times$ & $1.82\times$ & $2.16\times$ \\
        \bottomrule
    \end{tabular}
    \caption{Speedup over dense version when running sparsified OPT-2.7 models in DeepSparse.}
    \label{tab:cpu-inference}
\end{table}

\paragraph{GPU Speedups.} 2:4 sparsity as supported by NVIDIA GPUs of generation Ampere and newer theoretically offers $2\times$ acceleration of matrix multiplications. We now evaluate how big those speedups are in practice for the matmul problem sizes that occur in our specific models of interest. We use NVIDIA's official CUTLASS library (selecting the optimal kernel configuration returned by the corresponding profiler) and compare against the highly optimized dense cuBLAS numbers (also used by PyTorch). We assume a batch-size of 2048 tokens and benchmark the three matrix shapes that occur in OPT-175B; the results are shown in Table \ref{tab:gpu-speedups}. We measure very respectable speedups through 2:4 sparsity between $54 - 79\%$, for individual layers (end-to-end speedups will likely be slightly lower due to some extra overheads from e.g. attention).

\begin{table}[h!]
    \centering
    \begin{tabular}{|l|c|c|c|}
        \toprule
        Weight & Q/K/V/Out & FC1 & FC2 \\
        \midrule
        Dense & 2.84ms & 10.26ms & 10.23ms \\
        2:4 Sparse & 1.59ms & 6.15ms &6.64ms \\
        \midrule
        Speedup & $1.79\times$ & $1.67\times$ & $1.54\times$ \\
        \bottomrule
    \end{tabular}
    \caption{Runtime and speedup for the different layer shapes occuring in OPT-175B using 2048 tokens.}
    \label{tab:gpu-speedups}
\end{table}


\end{document}